\documentclass[fleqn, 10pt]{wlscirep}
\usepackage[utf8]{inputenc}
\usepackage[T1]{fontenc}
\usepackage{colortbl}
\usepackage{bm}
\usepackage{marvosym}

\title{PathBench: A comprehensive comparison benchmark for pathology foundation models
towards precision oncology}
\author[1,*]{Jiabo Ma}
\author[1,*]{Yingxue Xu}
\author[1,*]{Fengtao Zhou}
\author[1,+]{Yihui Wang}
\author[1,+]{Cheng Jin}
\author[1,+]{Zhengrui Guo}

\author[2]{Jianfeng Wu} % Wang Zhe group
\author[3]{On Ki Tang}  % Ronald CK Chan
\author[1]{Huajun Zhou} % HKUST
\author[1]{Xi Wang} % HKUST
\author[1,4]{Luyang Luo} % HKUST
\author[5]{Zhengyu Zhang} % Liang Li
\author[6,7,8]{Du Cai} % Gao Feng

\author[2]{Zizhao Gao} % Wang Zhe
\author[9]{Wei Wang} % Wang Zhe
\author[10]{Yueping Liu}    % hebeisiyuan
\author[10]{Jiankun He}     % hebeisiyuan
\author[11,12,13,14]{Jing Cui}  % qianfoshan
\author[15]{Zhenhui Li}
\author[16]{Jing Zhang}
\author[6,7,8, \Letter]{Feng Gao}
\author[16, \Letter]{Xiuming Zhang}
\author[5,17,18 \Letter]{Li Liang}
\author[3, \Letter]{Ronald Cheong Kin Chan}
\author[2,3, \Letter]{Zhe Wang}
\author[1,19,20,21,22, \Letter]{Hao Chen}

\affil[1]{Department of Computer Science and Engineering, Hong Kong University of Science and Technology, Hong Kong, China}
\affil[2]{State Key Laboratory of Holistic Integrative Management of Gastrointestinal Cancers, Department of Pathology, School of Basic Medicine and Xijing Hospital, Fourth Military Medical University, Xi’an, China}
\affil[3]{Department of Anatomical and Cellular Pathology, Chinese University of Hong Kong, Hong Kong, China}
\affil[4]{Department of Biomedical Informatics, Harvard University, Boston, USA.}
\affil[5]{Department of Pathology, Nanfang Hospital and School of Basic Medical Sciences, Southern Medical University, Guangzhou, China.}
\affil[6]{Department of General Surgery (Colorectal Surgery), The Sixth Affiliated Hospital, Sun Yat-sen University, Guangzhou, China}
\affil[7]{Guangdong Provincial Key Laboratory of Colorectal and Pelvic Floor Diseases, The Sixth Affiliated Hospital, Sun Yat-sen University, Guangzhou, China }
\affil[8]{Biomedical Innovation Center, The Sixth Affiliated Hospital, Sun Yat-sen University, Guangzhou, China}

\affil[9]{Department of Pathology, The First Affiliated Hospital of USTC, Division of Life Sciences and Medicine, University of Science and Technology of China, Hefei, China}
\affil[10]{Department of Pathology, The Fourth Hospital of Hebei Medical University, Shijiazhuang, China}
\affil[11]{Department of Pathology, The First Affiliated Hospital of Shandong First Medical University \& Shandong Provincial Qianfoshan Hospital, Ji'nan, China}
\affil[12]{Shandong medicine and Health Key Laboratory of Clinical Pathology, Ji'nan, China}
\affil[13]{Shandong Lung Cancer Institute, Ji'nan, China}
\affil[14]{Shandong Institute of Nephrology, Ji'nan, China}
\affil[15]{Department of Radiology, The Third Affiliated Hospital of Kunming Medical University,
Yunnan Cancer Hospital, Kunming, China.}
\affil[16]{Department of Pathology, The First Affiliated Hospital, School of Medicine, Zhejiang
University, Hangzhou, China}
\affil[17]{Guangdong Provincial Key Laboratory of Molecular Tumor Pathology, Guangzhou, China.}
\affil[18]{Jinfeng Laboratory, Chongqing, China}

\affil[19]{Department of Chemical and Biological Engineering, Hong Kong University of Science and Technology, Hong Kong, China}
\affil[20]{Division of Life Science, Hong Kong University of Science and Technology, Hong Kong, China}
\affil[21]{State Key Laboratory of Molecular Neuroscience, The Hong Kong University of Science and Technology, Hong Kong, China}
\affil[22]{HKUST Shenzhen-Hong Kong Collaborative Innovation Research Institute, Futian, Shenzhen, China}

\affil[*]{Contributed Equally (Co-first)}
\affil[+]{Contributed Equally (Co-second)}
\affil[\Letter]{Co-senior Authors}
\affil[ ]{\textbf{Lead Contact: Hao Chen  (jhc@cse.ust.hk)}}

\begin{abstract}
    The emergence of pathology foundation models has revolutionized computational
    histopathology, enabling highly accurate, generalized whole-slide image analysis
    for improved cancer diagnosis, treatment planning, and prognosis assessment.
    While these models show remarkable potential across cancer diagnostics and
    prognostics, their clinical translation faces critical challenges including variability
    in optimal model across cancer types, potential data leakage in evaluation, and
    lack of standardized benchmarks. Without rigorous, unbiased evaluation, even
    the most advanced PFMs risk remaining confined to research settings, delaying
    their life-saving applications. Existing benchmarking efforts remain limited
    by narrow cancer-type focus, potential pretraining data overlaps, or
    incomplete task coverage. 
    We present \textbf{PathBench}, the first comprehensive benchmark addressing these gaps through: multi-center in-hourse datasets spanning common cancers with rigorous leakage prevention, evaluation across the full clinical spectrum from diagnosis to prognosis, and an automated leaderboard system for continuous model assessment. 
    Our framework incorporates large-scale, clinically diverse data with standardized evaluation protocols, enabling objective comparison of PFMs while reflecting real-world clinical complexity. 
    All evaluation data comes from private medical providers, with strict exclusion of any pretraining usage to avoid data leakage risks.
    We have collected 15,888 whole-slide images (WSIs) from 8,549 patients across 10 hospitals, encompassing over 64 diagnosis and prognosis tasks. 
    Currently, our evaluation of 19 PFMs shows that Virchow2 and H-Optimus-1 are the most effective models overall.
    PathBench's dynamic
    \href{https://birkhoffkiki.github.io/PathBench/}{benchmark} supports ongoing
    community contributions through an automated evaluation pipeline. This work provides
    researchers with a robust platform for model development and offers clinicians
    actionable insights into PFM performance across diverse clinical scenarios,
    ultimately accelerating the translation of these transformative technologies
    into routine pathology practice.
\end{abstract}
\begin{document}
    \flushbottom
    \maketitle

    \thispagestyle{empty}

    \section*{Introduction}
    Histopathology serves as the cornerstone of modern oncology, guiding
    critical decisions from diagnosis to therapeutic strategy selection
    \cite{lu2021ai, song2023artificial, skrede2020deep}. While convolutional neural
    networks (CNNs) and vision transformers (ViTs) have demonstrated remarkable success
    in computational pathology through supervised learning \cite{shmatko2022artificial,
    verma2024sexually, shen2022explainable}, the field now stands at an
    inflection point with the rise of pathology foundation models (PFMs). These pre-trained
    models leverage self-supervised training on massive amounts of pathological images
    to learn powerful visual representations \cite{oquab2023dinov2, zhou2023cross,
    chen2021empirical}, or employ contrastive learning to align images, text,
    and even genetic information to further enhance the model's multimodal
    capabilities \cite{radford2021learning,yu2022coca, xu2024multimodal}. By pretraining
    on large-scale diverse data, PFMs are revolutionizing whole-slide image (WSI)
    analysis through three key advantages: superior generalization across institutions
    and staining protocols, reduced reliance on expensive extensive manual
    annotations, and emergent capabilities for multimodal reasoning in diagnostic
    contexts. The clinical potential of PFMs is evidenced by recent
    breakthroughs across multiple cancer types, including gastric inflammation \cite{bilal2025foundation},
    gastrointestinal cancer \cite{wang2025foundation}, breast cancer \cite{fournier2025extended},
    and other malignancies \cite{kondepudi2025foundation, mulliqi2025foundation}.
    Notably, these models have demonstrated proficiency not only in
    classification tasks but also in predicting molecular subtypes, treatment responses,
    and patient outcomes directly from histomorphological patterns.

    Despite these advances, three critical challenges hinder clinical translation
    of PFMs. First, optimal architecture and pretraining strategies show significant
    variability across cancer subtypes and clinical applications. Second, evaluation
    methodologies may suffer from data leakage or selection bias, particularly
    when test datasets overlap with pretraining data or share similar
    demographic characteristics. Third, the absence of standardized benchmarks
    makes it difficult to validate performance claims across real-world clinical
    settings. These challenges collectively underscore the need for a rigorous,
    leakage-free, and sustainable evaluation framework capable of objectively comparing PFM
    performance across diverse cancer types and clinical workflows.

    While existing benchmarking efforts have made valuable contributions, they face
    notable limitations. Many rely exclusively on public datasets that may not
    reflect clinical diversity and often contain hidden overlaps with model
    pretraining data
    \cite{neidlinger2024benchmarking,lee2025benchmarking,majzoub2025good}. Others
    focus narrowly on specific cancer types like prostate \cite{gustafsson2024evaluating}
    or ovarian cancer \cite{breen2025comprehensive}, limiting their
    generalizability. Even the most comprehensive studies \cite{campanella2025clinical}
    typically evaluate only a subset of clinically relevant tasks, neglecting
    critical aspects such as prognosis prediction and other multimodal tasks.

    To address these gaps, we present \textbf{PathBench}, the first comprehensive
    benchmark for PFMs in clinical data across common cancers. PathBench is
    designed to evaluate PFM performance on a wide range of tasks---from
    diagnosis to prognosis---using large-scale, multi-center datasets that reflect
    the diversity and complexity of real-world clinical scenarios. The benchmark data are obtained solely from private medical institutions, and rigorous protocols are employed to guarantee that none of the data had been exposed to evaluated PFMs during pretraining, thereby eliminating any risk of data contamination. Given the
    rapid advancement of PFMs and the growing need for broader cancer type
    coverage, we also establish a live leaderboard, hosted on our
    \href{https://birkhoffkiki.github.io/PathBench/}{GitHub} repository, to streamline
    the evaluation of new models and datasets. Model developers can submit their
    models and corresponding weights via pull requests, after which our standardized
    evaluation pipeline automatically assesses performance on in-house data and updates
    the leaderboard accordingly. By providing a unified evaluation framework and
    a dynamic leaderboard, PathBench aims to accelerate the development and validation
    of PFMs, enhancing their reliability and clinical applicability. This benchmark
    not only enables researchers to compare PFM performance across multiple cancer
    types and tasks but also serves as a critical resource for clinicians and pathologists
    to assess the real-world utility of these models in clinical practice.

    \section*{Results}
\begin{figure}[htbp]
        \centering
        \includegraphics[width=\textwidth]{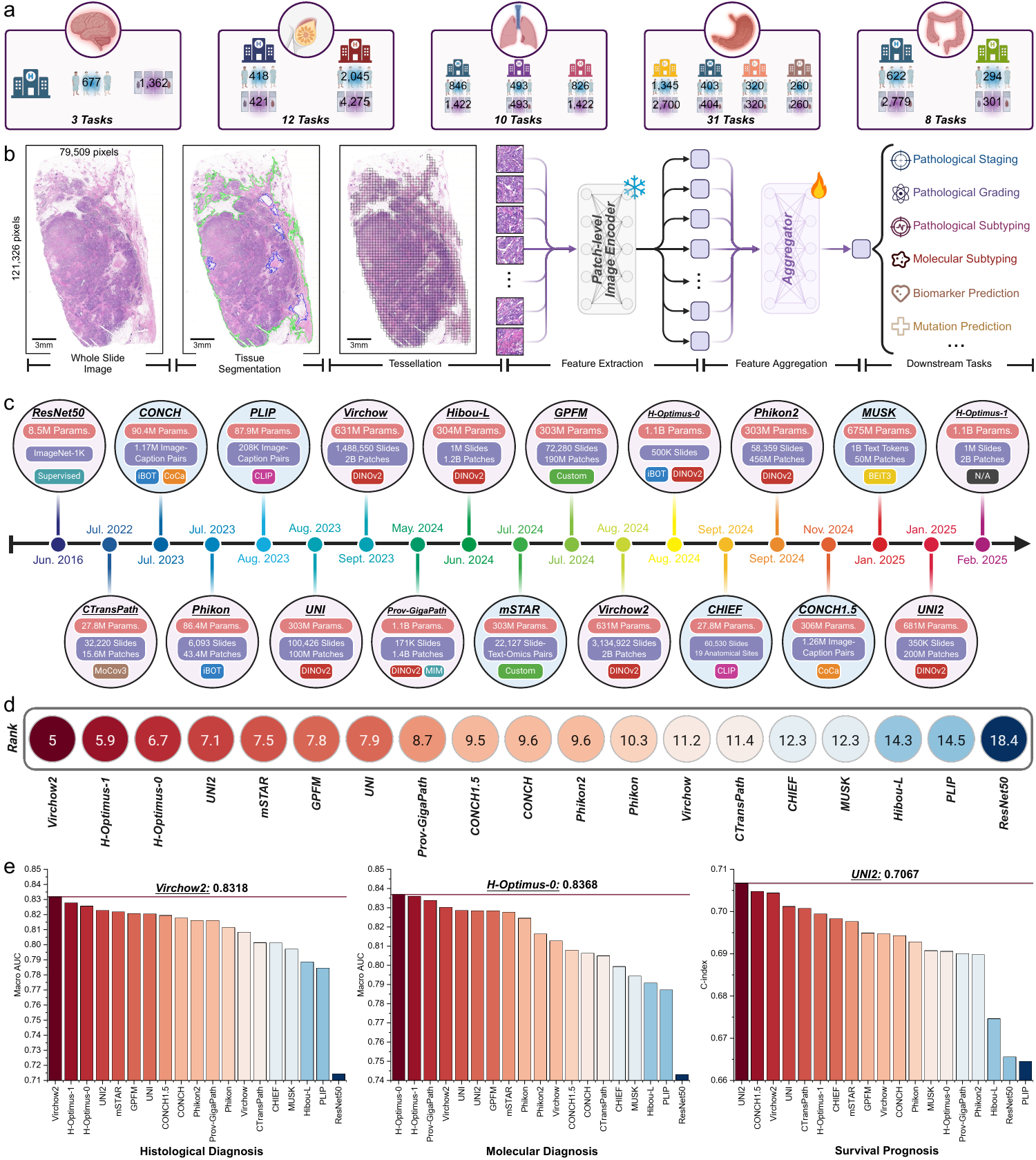}
        \caption{\textbf{The workflow and overall results of PathBench.} a. The
        data used for the benchmark. b. The workflow of evaluating foundation
        models. c. The evaluated foundation models. d. The average ranking score
        of evaluated models. f. The average performance on the pathological
        diagnosis, molecular diagnosis, and prognosis tasks. 
        }
        \label{Fig:main}
    \end{figure}

    We evaluated 19 pathology foundation models, including vision-only, vision-language, and multimodal-enhanced architectures (Figure \ref{Fig:main}c). 
    To ensure a comprehensive assessment, we tested these models on 64 tasks (Figure \ref{Fig:main}a) across five major cancer types: lung cancer
    (10 tasks), breast cancer (12 tasks), gastric cancer (31 tasks), colorectal cancer (8 tasks), and brain cancer (3 tasks). 
    For each cancer type, we conducted extensive experiments on clinically relevant tasks, including diagnosis, staging,
    molecular subtyping, biomarker prediction, and survival analysis (Figure \ref{Fig:main}b). 
    The evaluation utilized both internal validation sets and external cohorts to assess model generalizability and robustness across different clinical settings. 
    Among the 64 tasks, Virchow2, H-optimus-1, H-optimus-0, UNI2, and mSTAR achieved Top-5 performance, with
    rank scores of 5.0, 5.9, 6.6, 7.1, and 7.4, respectively (Figure \ref{Fig:main}d). 
    Specifically, Virchow2 and H-optimus-1 achieved Top-2 performance in histological subtyping tasks,
    while H-optimus-0 and H-optimus-1 excelled in molecular subtyping tasks. 
    UNI2 and CONCH1.5 demonstrated the best performance in survival prognosis tasks (Figure \ref{Fig:main}e). 
    In addition to overall performance, we also examined the models' effectiveness across different organs. Notably, H-optimus-1 ranked first in lung and colorectal cancer data, whereas Virchow2 performed best in breast, brain, and gastric cancer.
    Overall, the vision foundation models (e.g., Virchow2 and H-Optimus-1) are still more effective than vision-language models for the clinical-level tasks.

    \subsection*{Lung Cancer}
\begin{figure}
        \centering
        \includegraphics[width=\textwidth]{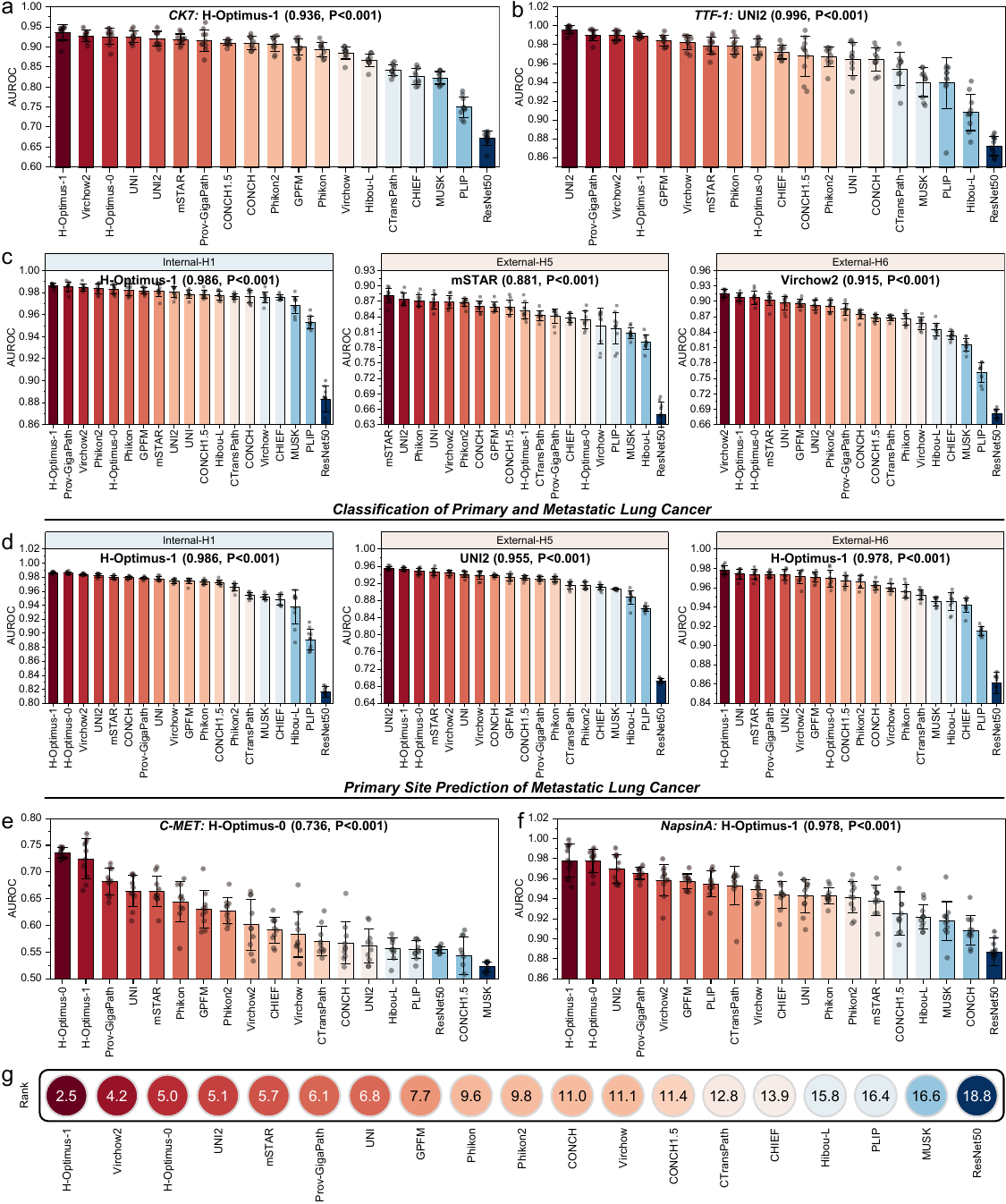}
        \caption{\textbf{Performance on lung cancer data.} a-b. Performance on
        molecular classification tasks CK7 and TTF-1. 
        c. Classification performance
        for primary LUAD versus metastatic lung cancer. 
        d. Primary site prediction
        for metastatic lung cancer. 
        e-f. Performance on molecular classification
        tasks C-MET and NpsinA.
        g. Overall ranking
        scores of foundation models on lung cancer data.
        The error bar indicates the standard deviation.
        }
        \label{Fig:Lung}
    \end{figure}

    Lung cancer is the leading cause of cancer-related deaths worldwide \cite{cao2024comparative}.
    In our evaluation of lung cancer data, we assessed the models on 10 tasks,
    including the classification of primary adenocarcinoma versus metastatic
    cancer, primary site prediction, and four molecular subtyping tasks based on
    H\&E slides. Overall, H-optimus-1 achieved the highest average ranking score
    of 2.5, followed by Virchow2 with a score of 4.2 (Figure \ref{Fig:Lung}g). For
    the metastatic cancer classification task on the internal cohort, all
    pathology foundation models performed well, with an AUC around 0.97;
    Virchow2 achieved the highest AUC of 0.9865. In two external cohorts, mSTAR
    and Virchow2 demonstrated the best performance, with AUCs of 0.8811 and 0.9152,
    respectively (Figure \ref{Fig:Lung}c). To predict the primary site of lung
    cancer, H-optimus-1 excelled, achieving AUCs of 0.9782 on the internal
    cohort and 0.9861 on the external cohort H6 (Figure \ref{Fig:Lung}d).
    Overall, both H-optimus-1 and Virchow2 are the top models for metastatic-related
    tasks. We also investigated the performance of different models on molecular
    subtyping tasks. The H-optimus series performed best on three out of four molecular
    subtyping tasks: CK7 (Figure \ref{Fig:Lung}a), C-MET (Figure \ref{Fig:Lung}e),
    and NapsinA (Figure \ref{Fig:Lung}f). 
    Predicting CK7 and NapsinA status
    proved relatively straightforward, with H-optimus-1 achieving AUCs of 0.9362
    and 0.9781, respectively. In contrast, predicting C-MET status was more
    challenging; the best-performing model, H-optimus, achieved an AUC of only 0.7362.
    For the TTF-1 task, UNI2 performed best with an AUC of 0.996 (Figure
    \ref{Fig:Lung}b), indicating that TTF-1 status prediction is relatively easy.
    In summary, for the lung cancer dataset, H-optimus-1 and Virchow2 are the
    optimal choices for clinical research and applications.

    \subsection*{Breast Cancer}
\begin{figure}[htbp]
        \centering
        \includegraphics[width=\textwidth]{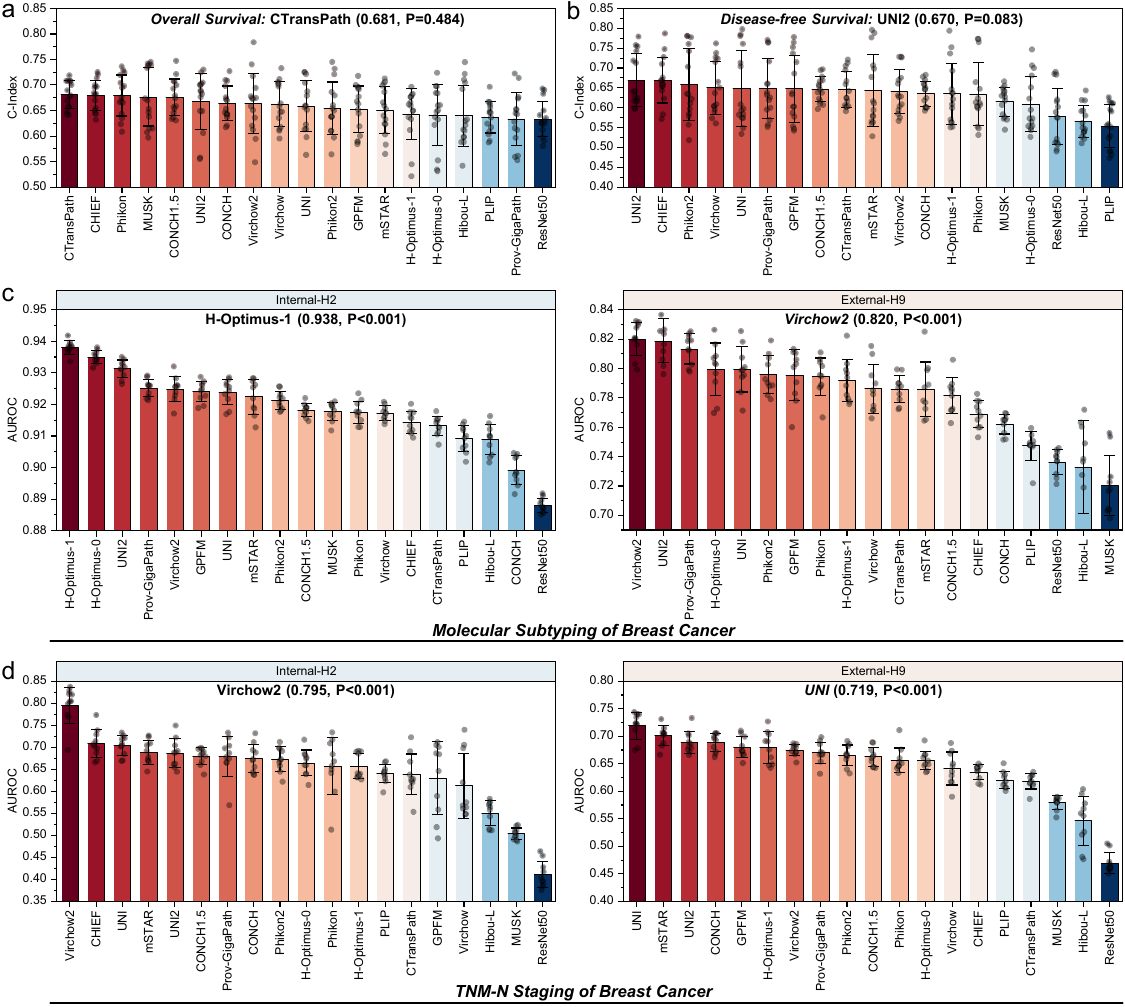}
        \caption{\textbf{The overall results of foundation models on the breast
        cancer data.} a.  Overall survival analysis
        results. b. Disease-free survival analysis results. c. Molecular
        subtyping results. d. TNM N Staging prediction results. 
        The error bars represent the standard deviation.
        }
        \label{Fig:Breast_1}
    \end{figure}

\begin{figure}[htbp]
        \centering
        \includegraphics[width=\textwidth]{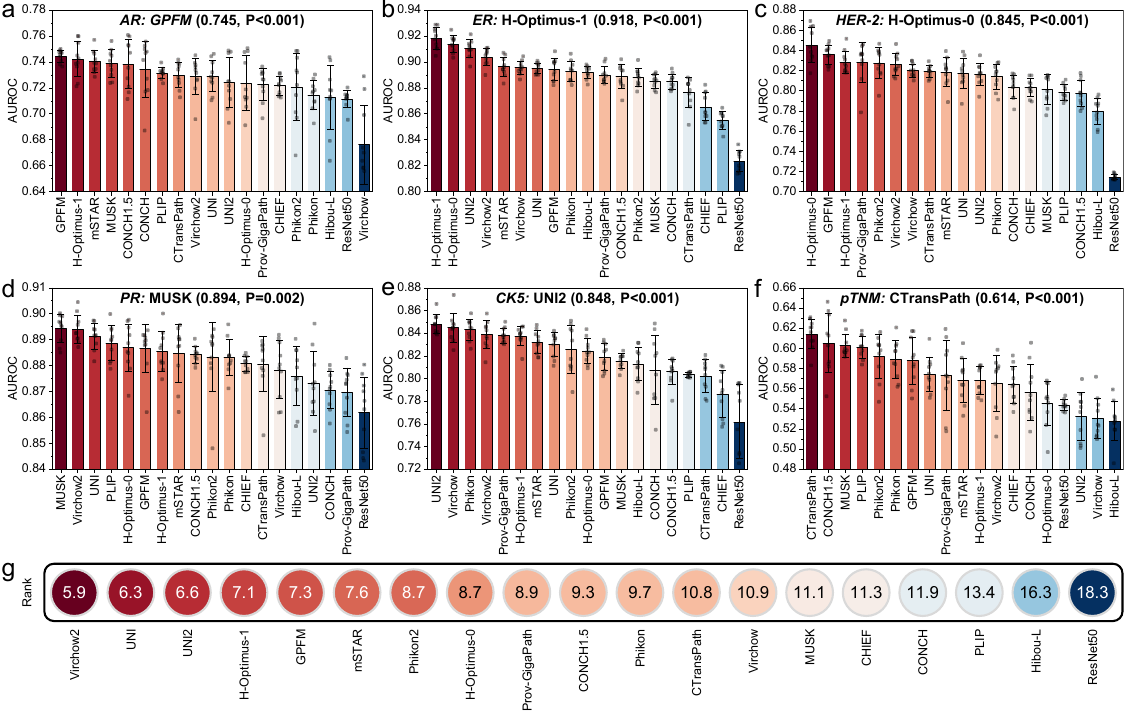}
        \caption{\textbf{The overall results of foundation models on the breast
        cancer data.} a-e. The molecular subtyping performance. f. pTNM Staging
        prediction results.
        g. The average ranking score of various foundation models on breast cancer tasks.
        The error bars represent the standard deviation.
        }
        \label{Fig:Breast_2}
    \end{figure}

    Breast cancer is the most prevalent cancer among women globally. The breast
    cancer dataset comprises 2,463 patients (4,696 WSIs) from two hospitals and
    covers 12 tasks, including molecular classification, subtype classification,
    and survival prediction. Results are presented in Figures \ref{Fig:Breast_1}
    and \ref{Fig:Breast_2}. 
Overall, Virchow2 demonstrates the best performance across most tasks, with an average rank of 5.9, closely followed by UNI at an average rank of 6.3 (Figure \ref{Fig:Breast_2}g).

    In molecular classification tasks, H-Optimus-1 achieves the highest
    performance on the internal cohort, with an AUC of 0.938. Conversely, Virchow2
    excels in the external cohort, recording an AUC of 0.8202 (Figure \ref{Fig:Breast_1}c).
    For diagnostic tasks, we evaluated the performance of foundation models on the
    TNM N staging task. Results indicate that Virchow2 leads with an accuracy of
    0.7949 on the internal cohort, while UNI performs best on the external cohort
    (Figure \ref{Fig:Breast_1}d). 
    These findings suggest that predicting TNM N stage from WSIs remains challenging. Additionally, in the pTNM staging task, the top-performing model, CTransPath, achieves an AUC of only 0.6142, indicating
    further complexity (Figure \ref{Fig:Breast_2}f). We also assessed these models
    across five molecular subtyping tasks: AR, ER, PR, HER2, and CK5. These tasks
    present significant challenges, with no single method consistently
    outperforming the others. GPFM, H-Optimus-1, MUSK, H-Optimus-0, and UNI2 achieved
    the highest performance on AR, ER, PR, HER2, and CK5, with AUCs of 0.7447, 0.9185,
    0.8944, 0.8454, and 0.8481, respectively (Figure \ref{Fig:Breast_2}a-e). Furthermore,
    we examined the models' performance in prognosis tasks. CTransPath and UNI2
    achieved the best results for overall survival and disease-free survival
    analysis, with C-Index of 0.6809 and 0.6697, respectively (Figures \ref{Fig:Breast_1}a-b).

    In summary, no single model dominated all tasks in the breast cancer dataset,
    indicating significant opportunities for improvement in foundation models applied
    to breast cancer data.

\subsection*{Gastric Cancer}
\begin{figure}[htbp]
        \centering
        \includegraphics[width=\textwidth]{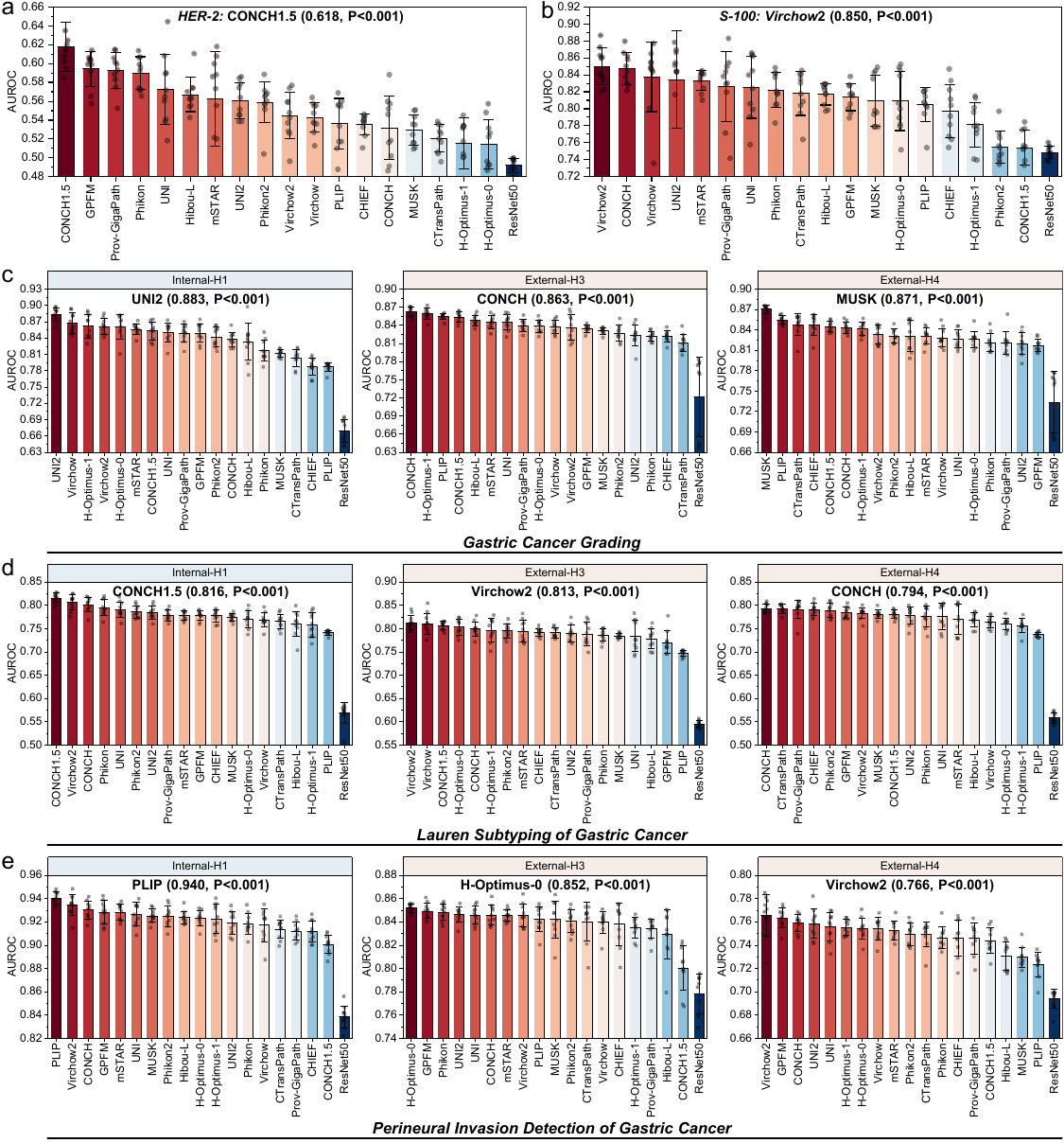}
        \caption{\textbf{Overall results of gastric cancer.} a-b. The molecular subtyping results of HER-2 and S-100, respectively.
        c. Gastric cancer grading results. 
        d. Lauren subtyping results. 
        e. Perineural invasion detection results. 
        The error bars represent the standard deviation.
        }
        \label{Fig:GC_1}
    \end{figure}

\begin{figure}[htbp]
        \centering
        \includegraphics[width=\textwidth]{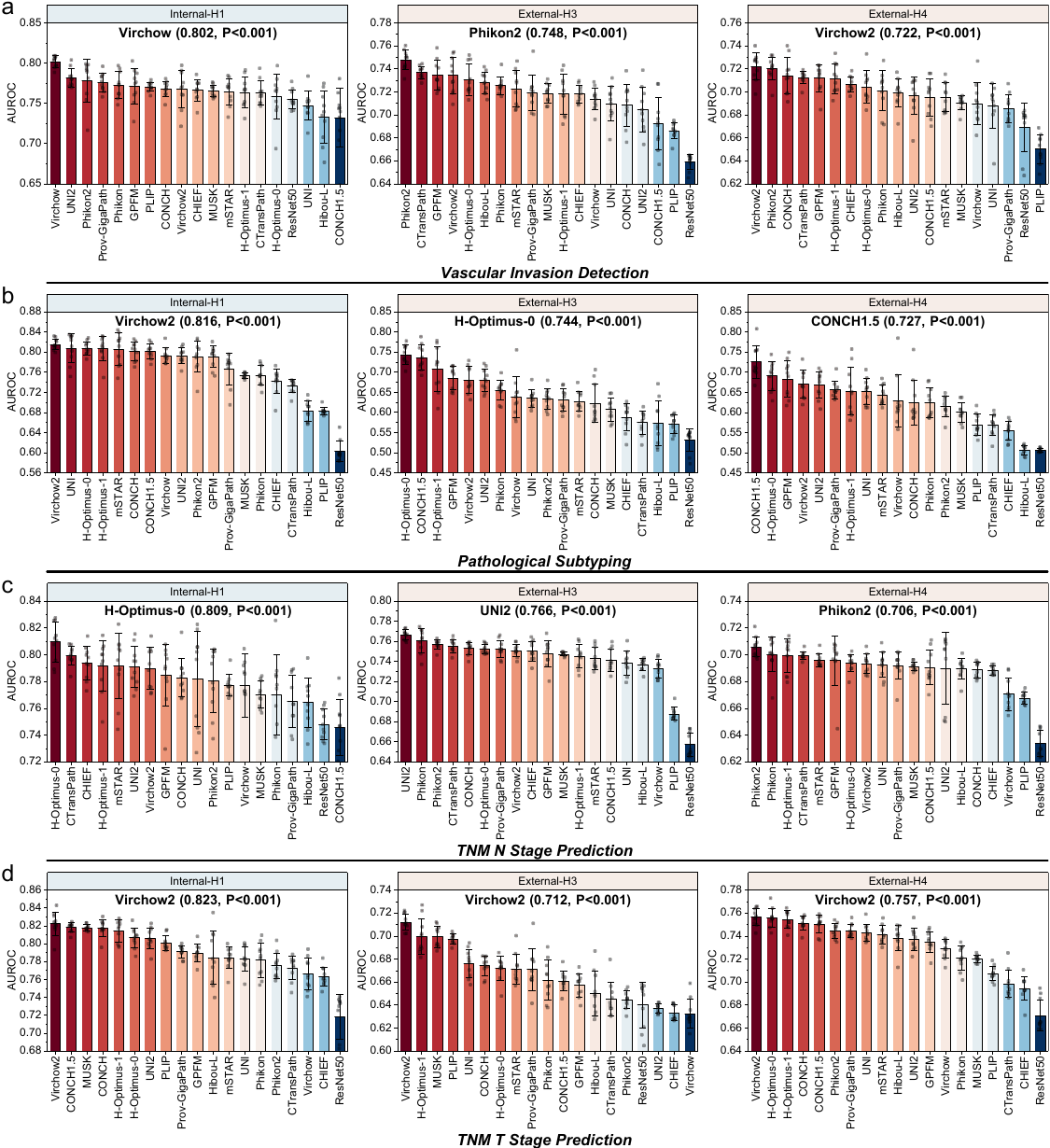}
        \caption{\textbf{Results of gastric cancer.} a. Vascular invasion
        detection results. b. Pathological subtyping results. 
        c. TNM N stage prediction results. 
        d. TNM T stage prediction results. 
        The error bars represent the standard deviation.
        }
        \label{Fig:GC_2}
    \end{figure}

\begin{figure}[htbp]
        \centering
        \includegraphics[width=\textwidth]{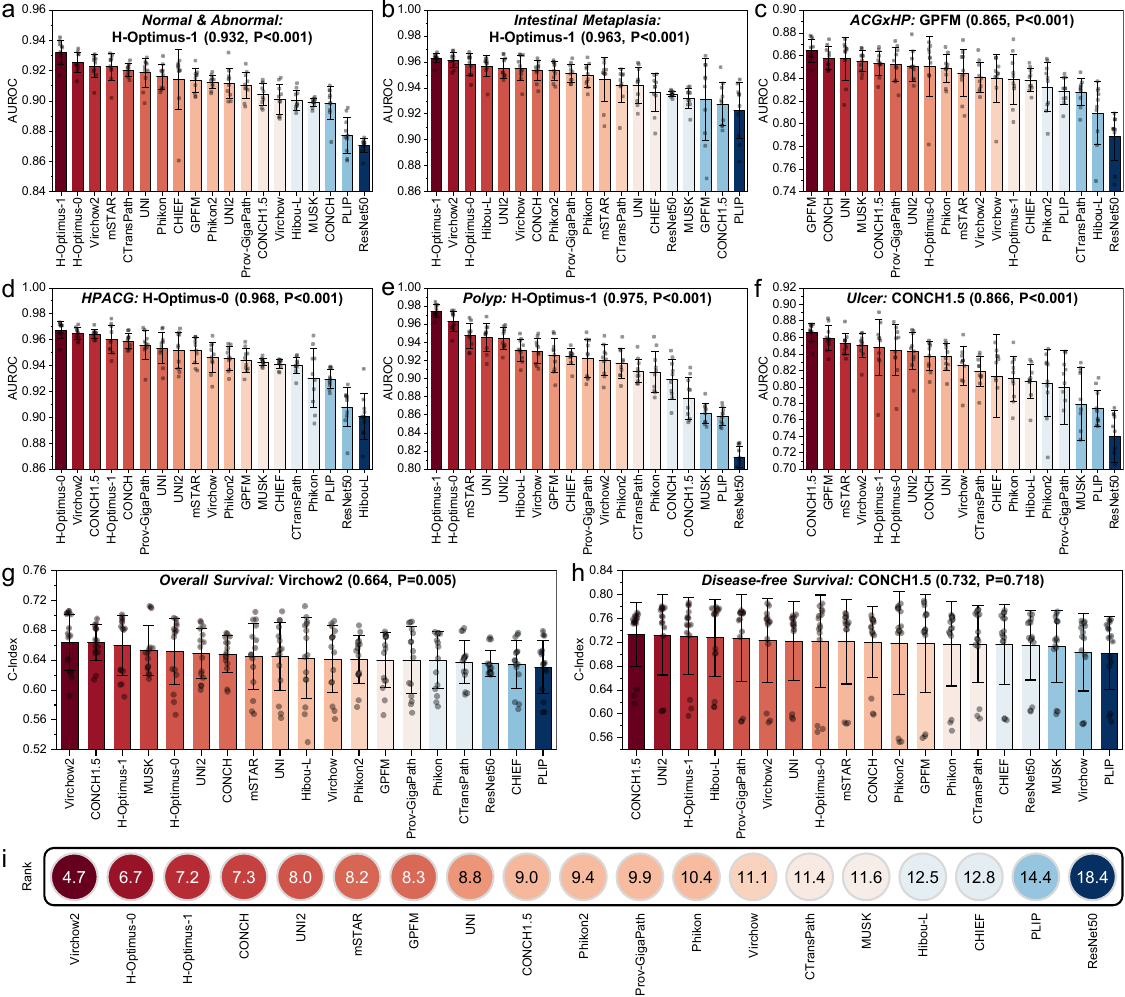}
        \caption{\textbf{Result of gastric cancer}.
        a. Normal and Abnormal slide classification.
        b. Detection of intestinal metaplasia.
        c. Classification of autoimmune chronic gastritis with Helicobacter pylori task.
        d. Detection of Helicobacter pylori-associated chronic gastritis.
        e-f. Detection of polyp an ulcer.
        g. The overall survival analysis results.
        h. The disease-free survival analysis results.
        i. The average ranking score of all foundation model on gastric cancer data.
               The error bars represent the standard deviation. 
        }
        \label{Fig:GC_3}
    \end{figure}

To evaluate the performance of foundation models on gastric cancer, we collected 3,684 WSIs from 2,328 patients across four hospitals, covering 31 tasks.
Overall, Virchow2 achieved the best performance, with an average ranking score of 4.7, followed by H-Optimus-0 at 6.7 (Figure \ref{Fig:GC_3}i).
We first assessed the models on pathological subtyping, Lauren subtyping, and gastric grade assessment tasks. Virchow2, CONCH1.5, and UNI2 demonstrated the highest performance with AUCs of 0.8159 (Figure \ref{Fig:GC_2}b), 0.8161 (Figure \ref{Fig:GC_1}d), and 0.8834 (Figure \ref{Fig:GC_1}c) on the internal cohort, respectively. 
Notably, vision-language models generally exhibited better generalizability on these tasks.
In addition, we evaluated the perineural and vascular invasion detection tasks.
Virchow2 and Phikon2 showed the best average performance on both internal and external datasets. However, all models performed poorly on the external dataset for the perineural invasion task (Figure \ref{Fig:GC_1}e). 
The vascular invasion detection task proved even more challenging, with no model achieving an AUC above 0.81 on both internal and external cohorts (Figure \ref{Fig:GC_2}a).
We further investigated the TNM staging tasks, specifically for N stage and T stage prediction. 
For the T stage, Virchow2 consistently performed best on the internal cohort (AUC of 0.8226) and two external cohorts (AUCs of 0.7122 and 0.7566) (Figure \ref{Fig:GC_2}d). 
For the N stage, H-Optimus, UNI2, and Phikon2 achieved the best performances on the internal and external cohorts H3 and H4, with AUCs of 0.8095, 0.7665, and 0.7095, respectively (Figure \ref{Fig:GC_2}c).
We also evaluated molecular subtyping tasks, where Virchow2 performed best  on the S-100 marker with an AUC of 0.8502. Predicting the HER-2 biomarker was more challenging, with the best-performing model, CONCH1.5, achieving only an AUC of 0.6179. Additionally, we assessed the models on prognosis tasks. 
Virchow2 and CONCH1.5 achieved the best performance in overall survival analysis and disease-free survival analysis, with C-Indexes of 0.664 and 0.7329, respectively (Figures \ref{Fig:GC_3}g-h).

Finally, we explored performance on gastric biopsy WSIs. H-Optimus-1 excelled in abnormal slide classification, intestinal metaplasia detection, and polyp detection, achieving AUCs of 0.9319, 0.9631, and 0.9746 respectively(Figure \ref{Fig:GC_3}a, \ref{Fig:GC_3}b and \ref{Fig:GC_3}e). 
For the Helicobacter pylori-associated chronic gastritis task, H-Optimus performed best with an AUC of 0.9676 (Figure \ref{Fig:GC_3}d). 
In the autoimmune chronic gastritis with Helicobacter pylori task, GPFM achieved the best performance with an AUC of 0.8645 (Figure \ref{Fig:GC_3}c). 
For ulcer detection, CONCH1.5 was the top model (Figure \ref{Fig:GC_3}f). 
Overall, H-Optimus-1 is the best choice for further research and applications in biopsy data.

\subsection*{Colorectal Cancer}
\begin{figure}[htbp]
        \centering
        \includegraphics[width=\textwidth]{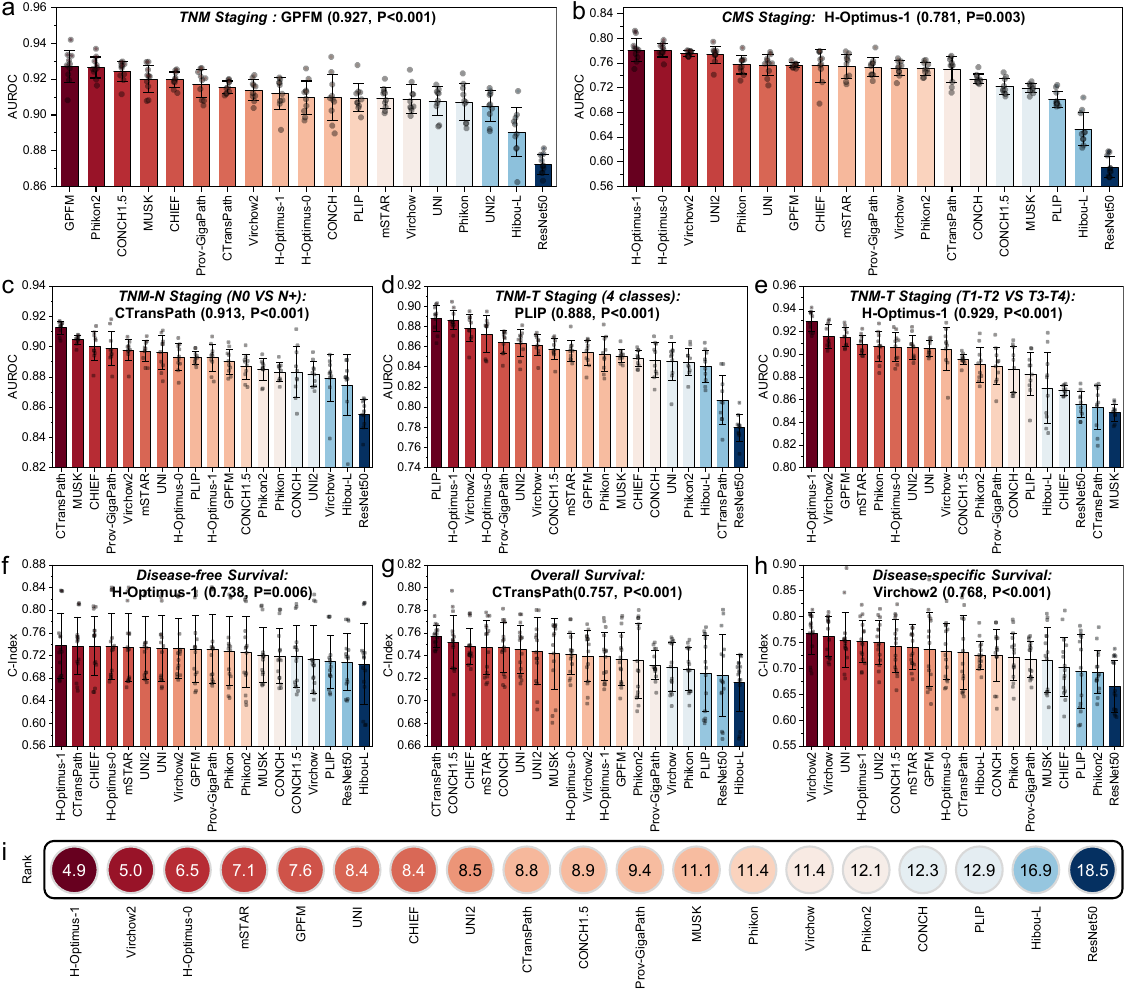}
        \caption{\textbf{Overall results on colorectal cancer data.}
        a-e. Staging tasks including TNM, CMS, TNM N, TNM T (4 classes), and TNM T (early or late stage).
        f-h. Survival analysis tasks including DFS, OS, and DSS.
         i. The average ranking score of the foundation model on the colorectal cancer data.
         The error bars represent the standard deviation. 
        }
        \label{Fig:Colon}
    \end{figure}

To evaluate foundation model on the colorectal cancer data, we collected 3,080 WSIs from 916 patients.
Overall, H-optimus-1 achieved the best performance with an average ranking score of 4.9, while Virchow2 ranked second with a score of 5.0 (Figure \ref{Fig:Colon}i).
We assessed these models on staging tasks, including TNM staging, further N staging, and T staging (both coarse-level and fine-level). 
No single model consistently outperformed the others across these four tasks (Figure \ref{Fig:Colon}a, \ref{Fig:Colon}c-e). 
Specifically, GPFM performed best on the TNM staging task with an AUC of 0.9272. 
For the further N staging task, CTransPath achieved the highest performance with an AUC of 0.9126. 
In the T staging task, H-optimus-1 excelled in coarse-level classification (2 classes) with an AUC of 0.9291, while PLIP attained the best performance with an AUC of 0.881 in fine-level classification (4 classes). 
Additionally, we evaluated the models on consensus molecular subtyping based on WSIs. The best-performing model, H-optimus-1, achieved an AUC of only 0.7814, indicating that consensus molecular subtyping is a more challenging task (Figure \ref{Fig:Colon}b).
We also reported the performance of foundation models on survival prognosis tasks, including overall survival (OS), disease-free survival (DFS), and disease-specific survival (DSS). 
Similarly, no model outperformed the others across all survival prognosis tasks (Figure \ref{Fig:Colon}f-h).
CTransPath, H-optimus-1, and Virchow2 excelled in OS, DFS, and DSS, with C-Index values of 0.7570, 0.7375, and 0.7678, respectively.
Overall, there is still room for improvement in the foundation models for colorectal cancer.

\subsection*{Brain Cancer}
\begin{figure}[htbp]
        \centering
        \includegraphics[width=\textwidth]{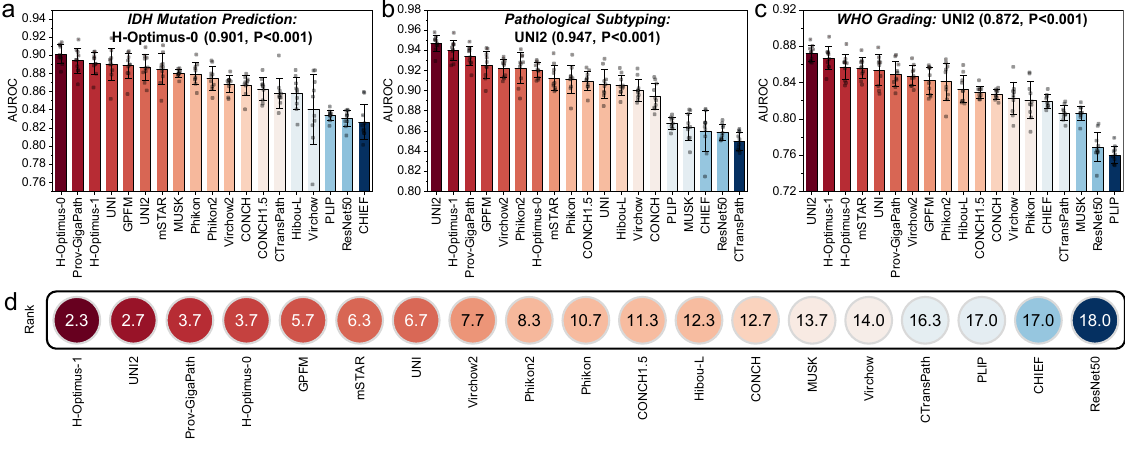}
        \caption{\textbf{Results on the brain cancer data.}
        a. Results of IDH Mutation prediction task.
        b. Performance on pathological subtyping task.
        c. Performance on the WHO Grading task.
         d. Average ranking score of various foundation models.
         The error bars represent the standard deviation. 
         }
        \label{Fig:Brain}
\end{figure}

We investigated foundation models on brain cancer data, focusing on IDH mutation prediction, pathological subtyping, and WHO grading tasks. 
To achieve this, we collected 1,362 slides from 677 patients. 
Overall, H-optimus-1 demonstrated the best performance with an average ranking score of 2.3, followed by UNI2 with a score of 2.7 (Figure \ref{Fig:Brain}d).
Specifically, H-optimus-1 excelled in the IDH mutation prediction task, achieving an AUC of 0.9013 (Figure \ref{Fig:Brain}a). 
In the pathological subtyping and WHO grading tasks, UNI2 outperformed with AUCs of 0.9469 and 0.8724, respectively (Figures \ref{Fig:Brain}b-c). 
Currently, both H-optimus-1 and UNI2 are strong performers in brain cancer analysis.

    \section*{Discussion}
    The comprehensive evaluation through PathBench reveals several critical insights into the current state and future directions of pathology foundation models (PFMs) in computational pathology. 
    Our benchmark demonstrates that while PFMs achieve remarkable performance across diverse diagnostic and prognostic tasks, their clinical applicability remains organ-specific and task-dependent.
    The observed performance variability across cancer types—with Virchow2 dominating gastric and breast cancer tasks, while H-Optimus-1 excels in lung and colorectal cancer.
    We highlight the following key findings.
    First, there is still significant room for improvement in vision-language models across most clinical tasks. 
    For example, the best performed vision-language model CONCH1.5 ranks 8 among all 19 models.
    Second, the consistent performance gap between internal and external validation cohorts underscores the importance of rigorous, leakage-free evaluation, as implemented in PathBench, prior to clinical deployment. 
    The results highlight the importance of improving the generalization ability of both the feature extractor and the aggregator. 
    Third, it is worth noting that simply increasing the number of whole slide images and the parameters of the model remains effective, as evidenced by the top two models (both trained on >1 million WSIs)..
    In addition, The strong performance of mSTAR—trained solely on TCGA data yet outperforming vision models trained on larger datasets—suggests that integrating multimodal knowledge can compensate for limited data.
    
    Our study has several limitations. While PathBench covers five major cancers, it does not yet address rare malignancies or pediatric tumors. Additionally, the interpretability of PFM decisions—a crucial factor for clinical adoption—remains unassessed. Future iterations of PathBench will incorporate explainability metrics and expand to include more cancer types through international collaborations.
    
    The PathBench framework establishes a critical foundation for translating PFMs into clinical practice. By maintaining a dynamic leader board with automated evaluation pipelines, we enable continuous benchmarking against evolving clinical standards. This approach addresses a key limitation of static benchmarks in fast-moving AI fields, while our leakage prevention protocols mitigate inflated performance claims. Moving forward, integration with real-world evidence platforms and prospective validation in diagnostic workflows will be essential to realizing the full potential of PFMs in precision oncology.

    \section*{Materials and Methods}
    \subsection*{1. Dataset}
    \subsubsection*{Lung Cancer}
    \textbf{(1) Primary Adenocarcinoma and Metastatic Cancer Classification}\\ Distinguishing
    primary lung adenocarcinoma from metastatic carcinoma is critical for determining
    appropriate treatment strategies. To evaluate the performance of existing
    foundation models on this classification task, we collected a dataset of 846
    cases from Hospital H1, comprising 389 primary cancers (686 WSIs) and 457
    metastatic cancers (736 WSIs). The data were label-stratified into training,
    validation, and test sets at a 7:1:2 ratio. To further validate model
    robustness, we incorporated two independent external cohorts:
    \begin{itemize}
        \item Hospital H5 cohort: 237 primary cases (237 WSIs) and 256
            metastatic cases (256 WSIs).

        \item Hospital H6 cohort: 465 primary cases (744 WSIs) and 361
            metastatic cases (678 WSIs).
    \end{itemize}
    \noindent
    \textbf{(2) Primary Site Prediction of Metastatic Lung Cancer}\\ For
    metastatic carcinomas, we curated an additional dataset from Hospital H1 to predict
    the primary tumor origin. This dataset comprised six primary sites: lung (393
    cases, 690 WSIs), colorectal (186 cases, 314 WSIs), kidney (25 cases, 36 WSIs), breast
    (55 cases, 80 WSIs), and liver (34 cases, 63 WSIs), with a 7:1:2 training-validation-test
    split. It is worth noting that the lung site only contains lung
    adenocarcinoma cases. Additionaly, we also collected data from Hospital H5
    and H6 for the external validation as follows:
    \begin{itemize}
        \item Hospital H5: Lung (237 cases, 237 WSIs), breast (50 cases, 50 WSIs),
            colorectal (96 cases, 96 WSIs), kidney (30 cases, 30 WSIs), and liver (10
            cases, 10 WSIs).

        \item Hospital H6: Lung (273 cases, 487 WSIs), colorectal (141 cases, 279
            WSIs), kidney (43 cases, 87 WSIs), breast (63 cases, 104 WSIs), and
            liver (5 cases, 10 WSIs).
    \end{itemize}
\noindent
    \textbf{(3) Biomarker Prediction of Lung Cancer}\\ Accurate prediction of
    biomarkers based on H\&E slides can help pathologists have a overall
    understanding of the tumor microenvironment and guide treatment decisions. To
    evaluate the performance of foundation models on biomarker prediction, we
    curated data from hospital H1 to perform the prediction of 4 biomarkers: C-MET,
    CK7, TTF-1, and Napsin A. The label is derived from the corresponding immunohistochemistry
    (IHC) results. The details of the dataset are as follows:\\
    \begin{itemize}
        \item \textbf{C-MET:} 195 C-MET negative slides, and 235 C-MET positive slides.

        \item \textbf{CK7:} 248 CK7 negative slides, and 171 CK7 positive slides.

        \item \textbf{TTF-1:} 369 TTF-1 negative slides, and 148 TTF-1 positive slides.

        \item \textbf{Napsin A:} 263 Napsin A negative slides, and 92 Napsin A positive
            slides.
    \end{itemize}
    The data were label-stratified into training, validation, and test sets at a
    7:1:2 ratio.

    \subsubsection*{Breast Cancer}
    \textbf{(1) TNM-N Staging of Breast Cancer}\\ Accurate lymph node staging (N
    stage) is crucial for breast cancer prognosis and treatment planning. To evaluate
    the performance of foundation models on this task, we curated a dataset from
    Hospital H2 and H9, focusing on N stage classification with two distinct categories:
    N0 (no regional lymph node metastasis) and N+ (presence of lymph node
    metastases, including both the N1 to N3 substages). The dataset from Hospital
    H2 includes 343 N0 cases (916 slides) and 125 N+ cases (381 slides). For training,
    validation and testing, the data are stratified at the case level in a 7:1:2
    ratio to avoid potential data leakage. In addition, we also collected data from
    Hospital H9 for external validation, which includes 62 N0 cases (62 slides)
    and 23 N+ cases (23 slides).

    \noindent
    \textbf{(2) pTNM Staging of Breast Cancer}\\ Furthermore, we established a
    dataset for evaluating foundation models in pTNM staging. The dataset are collected
    from Hospital H2, which includes 192 stage I cases (451 slides), 232 stage II
    cases (727 slides), and 43 stage III cases (116 slides). The data were case-level
    label-stratified into training, validation, and test sets at a 7:1:2 ratio,
    ensuring proportional representation of each stage across splits.

    \noindent
    \textbf{(3) Molecular Subtyping of Breast Cancer}\\ Accurate molecular
    subtyping of breast cancer is essential for personalized treatment
    strategies and prognostic assessment. To evaluate the performance of foundation
    models on this task, we curated datasets from Hospital H2 and H9. The dataset
    from Hospital H2 includes 307 Luminal A cases (310 slides), 614 Luminal B1
    cases (618 slides), 243 Luminal B2 cases (268 slides), 589 TNBC cases (1,932
    slides), and 292 HER-2 cases (323 slides). The data were case-level label-stratified
    into training, validation, and test sets at a 7:1:2 ratio. For external validation,
    we also collected data from Hospital H9, which includes 102 Luminal A cases,
    89 Luminal B1 cases, 24 Luminal B2 cases, 101 TNBC cases, and 102 HER-2 cases.
    Each cases contains only one slide.

    \noindent
    \textbf{(4) Biomarker Prediction of Breast Cancer}\\ Predicting biomarkers
    from H\&E slides can provide valuable insights and accelerate the diagnosis
    of breast cancer. To evaluate the performance of foundation models on biomarker
    prediction, we curated data from Hospital H2 to perform the prediction of 5
    biomarkers: AR, ER, PR, HER2, and CK5. The details of the dataset are as
    follows:
    \begin{itemize}
        \item \textbf{AR:} 463 AR negative cases (731 slides), and 677 AR positive
            cases (841 slides).

        \item \textbf{ER:} 767 ER negative cases (1,264 slides), and 781 ER positive
            cases (786 slides).

        \item \textbf{PR:} 623 PR negative cases (1,108 slides), and 933 PR positive
            cases (950 slides).

        \item \textbf{HER2:} 511 HER2 negative cases (743 slides), and 833 HER2 positive
            cases (975 slides).

        \item \textbf{CK5:} 753 CK5 negative cases (859 slides), and 208 CK5 positive
            cases (379 slides).
    \end{itemize}
    The data were case-level label-stratified into training, validation, and test
    sets at a 7:1:2 ratio, ensuring proportional representation of each
    biomarker across splits.

    \noindent
    \textbf{(5) Overall Survival Analysis of Breast Cancer}\\ Accurate
    prediction of overall survival (OS) is essential for guiding treatment
    decisions and prognostic stratification in breast cancer. We established a dataset
    for evaluating foundation models in predicting OS outcomes. The dataset are
    collected from Hospital H2 inlcuding 392 censored patients (1,089 slides) and
    59 deceased patients (181 slides). The 5-fold cross-validation was performed
    to evaluate the model performance. To further avoid the randomness, we also performed
    3 times of 5-fold cross-validation.

    \noindent
    \textbf{(6) Disease Free Survival Analysis of Breast Cancer}\\ Furthermore,
    we established a dataset for evaluating foundation models in disease-free
    survival (DFS) outcomes, where `recurred' is the outcome event. The dataset are collected from Hospital H2 inlcuding
    380 disease-free patients (1,066 slides) and 71 recurred patients
    (204 slides). The 5-fold cross-validation was performed to evaluate the
    model performance. To further avoid the randomness, we also performed 3
    times of 5-fold cross-validation.

    \subsubsection*{Gastric Cancer}
    \textbf{(1) Normal Gastric Biopsy Tissue and Abnormal tissue Classification}\\
    Distinguishing normal gastric bioposy tissue including chronic gastritis
    without Helicobacter pylori infection (normal/CGxHP) from abnormal tissue is
    essential for guiding clinical management and treatment decisions. To assess
    the performance of computational pathology foundation models on this diagnostic
    task, we compiled a dataset of 2,700 gastric biopsy slides from Hospital H7,
    comprising 733 normal/CGxHP slides and 1,967 abnormal slides. The data were
    label-stratified into training, validation, and test sets at a 7:1:2 ratio.

    \noindent
    \textbf{(2) Subtyping of Abnormal Gastric Biopsy Tissues}\\ For the abnormal
    gastric tissues, we further constructed a more practical multi-label classification
    task to evaluate the foundation models. The abnormal tissues from Hospital
    H7 contains 4 classes, including Helicobacter pylori-associated chronic gastritis
    (HPACG, 223 slides), Autoimmune chronic gastritis with Helicobacter pylori (ACGxHP,
    185 slides), Gastric polyps (144 slides), and Gastric ulcers (111 slides). Since
    one slide may correspond to multiple labels, we perform binary
    classification for each class. These slides were stratified by pathological labels
    into training (70\%), validation (10\%), and test sets (20\%).

    \noindent
    \textbf{(3) Binary Classification of Gastric Intestinal Metaplasia}\\
    Detecting intestinal metaplasia (IM) from non-IM gastric biopsy tissue is
    critical for early detection of precancerous lesions and risk stratification
    in gastric cancer screening. To evaluate the performance of foundation models
    on this diagnostic task, we curated a dataset of 2,700 gastric biopsy slides
    from Hospital H7, comprising 2,430 non-IM slides and 270 IM slides. The data
    were label-stratified into training, validation, and test sets at a 7:1:2
    ratio, ensuring proportional representation of IM cases across splits.

    % \noindent
    % \textbf{(4) Report Generation for Gastric Biopsy Slide}\\ Automated pathological
    % report generation can significantly improve diagnostic efficiency and reduce
    % inter-observer variability in clinical practice. To evaluate the capability
    % of foundation models in generating structured diagnostic reports, we curated
    % a dataset of 2,700 gastric biopsy slides from Hospital H7 with paired pathology
    % reports. These reports contained standardized descriptions of histological
    % features (e.g., inflammation severity, glandular architecture, Helicobacter pylori
    % status) and diagnostic conclusions. The dataset was stratified into training
    % (70\%), validation (10\%), and test sets (20\%), with careful preservation of
    % report templates and medical terminology across splits.

    \noindent
    \textbf{(4) Histopathological Grading Assesment of Gastric Cancer} \\
    Gastric cancer grading is a critical component of pathological assessment, reflecting
    the degree of tumor cell differentiation and correlating with biological behavior
    and prognosis. To evaluate foundation models for this task, we compiled a multi-institutional
    dataset. From Hospital H1, we included 318 poorly differentiated (G3) cases (319
    slides) and 81 well/moderately differentiated (G1+G2) cases (82 slides),
    with a 7:1:2 ratio split for training, validation, and testing. Additionally,
    we incorporated two additional cohorts for external validation: Hospital H3 contributed
    190 G3 cases and 55 G1+G2 cases, while Hospital H4 provided 258 G3 cases and
    62 G1+G2 cases.

    \noindent
    \textbf{(5) HER2 Status Prediction of Gastric Cancer}\\ Accurate HER2 status
    prediction is critical for guiding targeted therapy decisions in gastric
    cancer. To evaluate foundation models for this task, we compiled a multi-center
    dataset of 675 H\&E-stained slides from Hospitals H1, H3, and H4, with HER2 labels
    derived from corresponding Immunohistochemistry (IHC) results. The dataset
    includes 549 negative cases (IHC 0/1+) and 126 non-negative cases (IHC 2+/3+).
    Data were stratified into training (70\%), validation (10\%), and test sets
    (20\%) while preserving the original class distribution.

    \noindent
    \textbf{(6) S-100 Protein Expression Prediction of Gastric Cancer}\\ S-100 protein
    expression in gastric cancer has been associated with tumor differentiation,
    neuroendocrine differentiation, and potential prognostic implications. To
    evaluate the ability of computational pathology models in predicting S-100
    status from H\&E-stained slides, we compiled an multi-center dataset from Hospital
    H1 and H3. The dataset contains 90 IHC 0 slides and 270 IHC 1+ slides. The
    dataset were label-stratified into training, validation, and testing sets at
    a ratio of 7:1:2.

    \noindent
    \textbf{(7) Lauren classification of Gastric Cancer}\\ The Lauren classification
    system is pivotal in gastric cancer prognostication and therapeutic decision-making,
    categorizing tumors into distinct histomorphological subtypes with differing
    biological behaviors. To evaluate foundation models for this critical task,
    we constructed a multi-center dataset encompassing three Lauren subtypes: Diffuse
    type, Intestinal type, and Mixed type. For the internal cohort, the dataset comprised
    388 cases from Hospital H1 including 159 diffuse cases (160 slides), 102
    intestinal cases (103 slides), and 127 mixed cases (127 slides). All data
    were label-stratified into training (70\%), validation (10\%), and test sets
    (20\%), with subtype proportions preserved across splits. To ensure generalizability,
    two independent external datasets were included:
    \begin{itemize}
        \item Hospital H3: 141 slides (77 Diffuse, 33 Mixed, 31 Intestinal)

        \item Hospital H4: 319 slides (143 Diffuse, 86 Mixed, 90 Intestinal)
    \end{itemize}

    \noindent
    \textbf{(8) Pathological Subtyping of Gastric Cancer}\\ Accurate
    classification of gastric cancer pathological subtypes is essential for
    prognostic stratification and therapeutic planning. To evaluate foundation models
    on this task, we curated a dataset from Hospital H1 including 163 Signet
    Ring Cell Carcinoma (163 slides), 166 Tubular Adenocarcinoma (167 slides),
    and 66 non-specified Stomach Adenocarcinoma (67 slides). The internal data were
    stratified into training (70\%), validation (10\%), and test sets (20\%)
    while preserving subtype proportions. Additionally, to assess model
    generalizability, we also adopted data from Hospital H3 and H4 as the external
    validation.
    \begin{itemize}
        \item Hospital H3: 233 Stomach Adenocarcinoma, 82 Signet Ring Cell

        \item Hospital H4: 195 Stomach Adenocarcinoma, 59 Signet Ring Cell
    \end{itemize}

    \noindent
    \textbf{(9) Detection of Perineural Invasion in Gastric Cancer}\\ Perineural
    invasion (PNI) is an important prognostic factor in gastric cancer associated
    with increased recurrence risk and poor survival outcomes. To evaluate
    foundation models for this critical histopathological feature, we compiled a
    dataset with standardized PNI assessment from Hospital H1, consisting of 255
    PNI-positive cases (256 slides) and 141 PNI-negative cases (142 slides). The
    internal data were stratified into training (70\%), validation (10\%), and test
    sets (20\%) while maintaining the original PNI positivity rate. we also included
    two independent cohort from Hospital H3 and H4 as the external validation as
    follows:
    \begin{itemize}
        \item Hospital H3: 156 PNI-positive slides and 76 PNI-negative slides

        \item Hospital H4: 112 PNI-positive slides and 207 PNI-negative slides
    \end{itemize}

    \noindent
    \textbf{(10) Detection of Vascular Invasion in Gastric Cancer}\\ Vascular invasion
    (VI) is a critical histopathological feature in gastric cancer that correlates
    with hematogenous metastasis risk and guides adjuvant therapy decisions. To
    assess foundation models for VI identification, we established a dataset with
    balanced representation of positive and negative cases from Hospital H1. The
    internal cohort contained 197 VI-positive cases (198 slides) and 198 VI-negative
    cases (199 slides). To validate the robustness of evaluted models, we
    included another two cohort from Hospital H3 and H4 as the external
    validation as follows:
    \begin{itemize}
        \item Hospital H3: 90 VI-negative and 140 VI-positive slides

        \item Hospital H4: 197 VI-negative and 122 VI-positive slides
    \end{itemize}
    The internal data were stratified into training (70\%), validation (10\%), and
    test sets (20\%) while preserving the balanced class distribution.
    % External data were used exclusively for testing, with case-level splitting to prevent data leakage in multi-slide specimens.

    \noindent
    \textbf{(11) N Staging Classification in Gastric Cancer}\\ Accurate nodal
    status (N stage) determination is crucial for gastric cancer prognosis and
    treatment planning. We developed a clinically relevant binary classification
    task distinguishing node-negative (N0) from node-positive (N+, encompassing
    N1-N3) cases to evaluate foundation models' performance in this critical
    diagnostic task. The internal dataset comprised 212 N+ cases (212 slides)
    and 186 N0 cases (188 slides). For the external validation, we include two
    cohorts from Hospital H3 and H4 as follows:
    \begin{itemize}
        \item Hospital H3: 175 N+ slides and 85 N0 slides

        \item Hospital H4: 175 N+ slides and 145 N0 slides
    \end{itemize}

    All internal data were stratified into training (70\%), validation (10\%),
    and test sets (20\%) while preserving the original N0/N+ ratio.
    % External data were reserved exclusively for evaluating model generalizability across different institutional staging practices.

    \noindent
    \textbf{(12) T Staging Classification in Gastric Cancer}\\ Precise
    assessment of tumor invasion depth (T stage) is fundamental for gastric
    cancer treatment stratification and surgical planning. We evaluated foundation
    models on this critical four-class classification task (T1-T4) using a multi-institutional
    dataset. The internal dataset was collected from Hospital H1 comprising 115
    T1 slides (115 cases), 55 T2 slides (54 cases), 121 T3 slides (121 cases), and
    107 T4 slides (106 cases). To assess generalizability across different
    hospitals, we collected two cohorts from Hospital H3 and H4 for external validation
    as follows:
    \begin{itemize}
        \item Hospital H3: 260 slides (33 T1, 35 T2, 75 T3, 117 T4)

        \item Hospital H4: 320 slides (64 T1, 46 T2, 125 T3, 85 T4)
    \end{itemize}
    All internal data were stratified into training (70\%), validation (10\%),
    and test sets (20\%) while preserving T-stage proportions.
    % External cohorts were reserved exclusively for testing, with case-level splitting to prevent data leakage.

    \noindent
    \textbf{(13) Disease Free Survival Analysis of Gastric Cancer}\\ Prognostic prediction
    of disease recurrence is crucial for postoperative surveillance and adjuvant
    therapy planning in gastric cancer. We established a dataset for evaluating
    foundation models in predicting disease-free survival (DFS) outcomes based
    on WSIs. The dataset are collected from Hospital H3 comprising 260 slides from
    260 cases: 157 disease-free patients and 103 recurred
    patients. The 5-fold cross-validation was performed to evaluate the model performance.
    To further avoid the randomness, we also performed 3 times of 5-fold cross-validation.

    \noindent
    \textbf{(14) Overall Survival Analysis of Gastric Cancer}\\ Accurate prediction
    of overall survival (OS) is essential for guiding treatment decisions and prognostic
    stratification in gastric cancer. We established a dataset for evaluating
    foundation models in predicting OS outcomes. The dataset are collected from Hospital
    H3 comprising 260 slides from 260 cases: 172 censored
    patients and 88 uncensored patients. The 5-fold cross-validation was performed
    to evaluate the model performance. To further avoid the randomness, we also performed
    3 times of 5-fold cross-validation.

    \subsubsection*{Colorectal Cancer}
    \textbf{(1) Lymph Node Staging (N Stage) in Colorectal Cancer}\\ Lymph node metastasis
    in colorectal cancer, a critical component of TNM staging that significantly
    impacts treatment planning and outcome prediction. To evaluate the performance
    of foundation models in this task, we curated a data set from Hospital H8, focusing
    specifically on N-stage classification with two distinct categories: N0 (no regional
    lymph node metastasis) and N+ (presence of lymph node metastases, including both
    the N1 and N2 substages). The collection comprises 367 N0 cases (1,848
    slides) and 230 N+ cases (871 slides). For training, validation and testing,
    the data are stratified at the case level in a 7:1:2 ratio to avoid potential
    data leakage.

    \noindent
    \textbf{(2) Tumor Invasion Depth (T Stage) in Colorectal Cancer (2 classes)}\\ Tumor
    invasion depth (T stage) is a key determinant in colorectal cancer prognosis and treatment
    strategy, distinguishing between early-stage (T1+T2) and advanced local
    invasion (T3+T4). To assess foundation models' capability in this diagnostic
    task, we compiled the a dataset from Hospital H8, grouping cases into two
    clinically relevant categories: T1+T2 (tumor confined to bowel wall) and T3+T4
    (tumor extending beyond muscularis propria). It contains 519 T3+T4 cases (2,391
    slides) and 76 T1+T2 cases (319 slides). The data is rigorously case-level split
    into training (70\%), validation (10\%), and test sets (20\%) to ensure
    clinically meaningful evaluation.

    \noindent
    \textbf{(3) Tumor Invasion Depth (T Stage) in Colorectal Cancer (4 classes)} To further
    challenge the models' discriminative capabilities, we extended the task to fine-grained
    T-stage prediction (T1, T2, T3, T4). This 4-class classification problem reflects
    the full spectrum of tumor invasion depth, demanding higher precision from
    the models. The dataset includes 595 cases (2,710 slides) with the following
    distribution: 20 T1 (71 slides), 26 T2 (244 slides), 440 T3 (2,130 slides), and
    79 T4 (261 slides). Consistent with the 2-class task, we applied a case-level
    split (training:validation:testing = 7:1:2) to maintain evaluation integrity
    and prevent data leakage.

    \noindent
    \textbf{(4) TNM Staging of Colorectal Cancer}\\ Accurate TNM staging is fundamental
    for colorectal cancer management, directly guiding therapeutic decisions and
    prognostic evaluation. To evaluate foundation models in this task, we
    curated a dataset from Hospital H8 including 63 stage I cases (263 slides), 288
    stage II cases (1535 slides), 135 stage III cases (633 slides), and 120 stage
    IV cases (120 slides). The data were case-level stratified into training, validation,
    and testing at a ratio of 7:1:2.

    \noindent
    \textbf{(5) Consensus Molecular Subtyping in Colorectal Cancer}\\ Consensus molecular
    subtyping in colorectal cancer is crucial for precision oncology and treatment stratification.
    To evalute the performance of foundation models in this task, we curated a dataset
    from Hospital H8 containing 588 molecularly-characterized cases classified into
    four Consensus Molecular Subtypes (CMS): 76 CMS1 (372 slides), 239 CMS2 (1061
    slides), 86 CMS3 (393 slides), and 187 CMS4 (857 slides). For training, validation,
    and testing, the data were case-level stratified at a ratio of 7:1:2.

    \noindent
    \textbf{(6) Disease Free Survival Analysis of Colorectal Cancer}\\ Accurate
    prediction of disease-free survival (DFS) is crucial for postoperative
    surveillance and adjuvant therapy planning in colorectal cancer. We established a
    dataset for evaluating foundation models in predicting DFS outcomes based on
    WSIs. The dataset are collected from Hospital H8 comprising 2,779 slides
    from 608 cases. It contains 389 disease-free patients (1,874 slides) and 219
    recurred/progressed patients (904 slides). The 5-fold cross-validation was
    performed to evaluate the model performance. To further avoid the randomness,
    we also performed 3 times of 5-fold cross-validation.

    \noindent
    \textbf{(7) Disease Specific Survival Analysis of Colorectal Cancer}\\ We further
    established a dataset for evaluating foundation models in disease-specific
    survival (DSS) outcomes based on WSIs. The dataset are collected from Hospital
    H1 inlcuding 294 patients (301 slides). It contains 252 living, or dead but tumor-free patients (259 slides) and 42 dead patients with tumor (42 slides). The 5-fold
    cross-validation was performed to evaluate the model performance. To further
    avoid the randomness, we also performed 3 times of 5-fold cross-validation.

    \noindent
    \textbf{(8) Overall Survival Analysis of Colorectal Cancer}\\ Finally, we
    established a dataset for evaluating foundation models in overall survival (OS)
    outcomes based on WSIs. The dataset are collected from Hospital H8 inlcuding
    608 patients (2,779 slides). It contains 440 living patients (2,081 slides)
    and 168 deceased patients (698 slides). The 5-fold cross-validation was performed
    to evaluate the model performance. To further avoid the randomness, we also performed
    3 times of 5-fold cross-validation.

    \subsubsection*{Brain Cancer}
    \textbf{(1) Histopathological Subtyping of Glioma} Accurate classification
    of glioma subtypes is critical for precise diagnosis and treatment planning
    in neuro-oncology. To evaluate the performance of computational pathology models
    on this task, we constructed a dataset comprising 1,353 WSIs from 673 cases
    collected at Hospital H1. The dataset includes four major glioma subtypes:
    274 Glioblastoma cases (558 slides), 231 Diffuse astrocytoma cases (450
    slides), 132 Oligodendroglioma (269 slides), and 36 Diffuse midline glioma (76
    slides). The data were label-stratified into training (70\%), validation (10\%),
    and test sets (20\%).

    \noindent
    \textbf{(2) WHO Grading of Glioma} This task is designed for automated WHO
    grading of gliomas, a critical determinant of clinical management and prognosis
    in neuropathology. This dataset comprises 1,350 WSIs from 672 cases
    collected at Hospital H1, covering three WHO grades: 310 Grade 4 cases (634 slides),
    117 Grade 3 cases (234 slides), and 245 Grade 2 cases (482 slides). The dataset
    is lebel-stratified into training, validation, and testing at a ratio of 7:1:2.

    \noindent
    \textbf{(3) IDH Mutation Prediction of Glioma} Isocitrate dehydrogenase (IDH)
    mutation is a critical molecular marker with diagnostic, prognostic, and therapeutic
    implications in glioma management. To evaluate the potential of pathology
    foundation models in predicting mutation status from histopathological WSIs,
    we constructed a dataset comprising 1,341 slides from 667 cases collected from
    Hospital H1, categorized into two classes: 275 IDH-mutant cases (562 slides)
    and 392 IDH-wildtype cases (779 slides). The data is label stratified into a
    training, validation, and test sets at a ratio of 7:1:2.

    \subsection*{2. Model}

\begin{table}[htbp]
  \centering
  \caption{Pathology Foundation Models included in PathBench. MedIA, NM and NBME refer to Medical Image Analysis, Nature Medicine and Nature Biomedical Engineering, respectively. All parameter counts are based on actual measurements from the released model weights.}
  \scalebox{0.7}{
    \begin{tabular}{c|c|c|c|c|c|c|c|c|c}
    \toprule
    \textbf{PFM} & \textbf{\#Slides} & \textbf{\#Patches} & \textbf{Params.} & \textbf{Architecture} & \textbf{Pretraining Strategy} & \textbf{Pretraining Data Source} & \textbf{Stain} & \textbf{Released Date} & \textbf{Publication} \\
    \midrule
    \rowcolor[rgb]{ .949,  .949,  .949} \multicolumn{10}{c}{\textit{Vision-only}} \\
    \midrule
    ResNet50~\cite{r50} & -     & -     & 8.5M   & ResNet-50 & Supervised Learning & ImageNet-1K & -     & Jun-16 & CVPR \\
    \midrule
    CtransPath~\cite{ctranspath} & 32,220 & 15.6M & 27.8M & Swin-T/14 & MoCov3 & TCGA, PAIP & H\&E  & Jul-22 & MedIA \\
    \midrule
    Phikon~\cite{phikon} & 6,093 & 43.4M & 86.4M & ViT-B/16 & iBOT  & TCGA  & H\&E  & 26-Jul-23 & Preprint \\
    \midrule
    UNI~\cite{uni}   & 100,426 & 100M  & 303M  & ViT-L/16 & DINOv2 & GTEx, In-house & H\&E  & 29-Aug-23 & NM \\
    \midrule
    Virchow~\cite{virchow} & 1,488,550 & 2.0B  & 631M  & ViT-H/14 & DINOv2 & In-house & H\&E  & 14-Sep-23 & NM \\
    \midrule
    Prov-GigaPath~\cite{gigapath} & 171K  & 1.4B  & 1.1B  & ViT-G/14 & DINOv2, MIM & In-house & H\&E, IHC & 22-May-24 & Nature \\
    \midrule
    Hibou-L~\cite{hibou} & 1.1M    & 1.2B  & 304M  & ViT-L/14 & DINOv2 & -     & H\&E, others & 7-Jun-24 & Preprint \\
    \midrule
    GPFM~\cite{gpfm}  & 72,280 & 190M  & 303M  & ViT-L/14 & Custom & 33 Public datasets & H\&E  & 26-Jul-24 & NBME \\
    \midrule
    Virchow2~\cite{virchow2} & 3,134,922 & 2.0B  & 631M  & ViT-H/14 & DINOv2 & In-house & H\&E, IHC & 1-Aug-24 & Preprint \\
    \midrule
    H-Optimus-0~\cite{hoptimus0} & 500K  & -     & 1.1B  & ViT-G/14 & iBOT, DINOv2 & -     & -     & 8-Aug-24 & - \\
    \midrule
    Phikon2~\cite{phikon2} & 58,359 & 456M  & 303M  & ViT-L/16  & DINOv2 & 132 Public datasets & H\&E, IHC & 13-Sep-24 & Preprint \\
    \midrule
    UNI2~\cite{uni}  & 350K  & 200M  & 681M  & ViT-H/14 & DINOv2 & GTEx, In-house & H\&E, IHC & 14-Jan-25 & NM \\
    \midrule
    H-Optimus-1~\cite{hoptimus1} & 1M    & 2.0B  & 1.1B  & ViT-G/14 & -     & -     & -     & 1-Feb-25 & - \\
    \midrule
    \rowcolor[rgb]{ .949,  .949,  .949} \multicolumn{10}{c}{\textit{Vision-Language}} \\
    \midrule
    PLIP~\cite{plip}  & -     & 208K  & 87.9M   & ViT-B/32 & CLIP  & Tweets, Replies, LAION-5B & H\&E, IHC & 17-Aug-23 & NM \\
    \midrule
    CONCH~\cite{conch} & 21,442 & 1.17M   & 90.4M   & ViT/B-16 & iBOT, CoCa & PubMed, EDU, In-house & H\&E, IHC & 24-Jul-23 & NM \\
    \midrule
    CHIEF~\cite{chief} & 60,530 & -     & 27.8M & Swin-T/14 & CLIP  & \parbox{3cm}{8 public datasets, \\6 in-house datasets} & H\&E  & 4-Sep-24 & Nature \\
    \midrule
    CONCH1.5~\cite{conch} & -     & 1.26M & 306M  & ViT-L/16 & CoCa  & -     & H\&E, IHC & Nov-24 & - \\
    \midrule
    MUSK~\cite{musk}  & 33K   & 1M  & 675M  & ViT-L/16 & BEiT3 & \parbox{3cm}{\centering PubMed, TCGA, \\QUILT-1M, PathAsst} & H\&E  & 8-Jan-25 & Nature \\
    \midrule
    \rowcolor[rgb]{ .949,  .949,  .949} \multicolumn{10}{c}{\textit{Multimodal-enhanced}} \\
    \midrule
    mSTAR~\cite{mstar} & 22,127 & 116M  & 303M  & ViT-L/16 & Custom & TCGA  & H\&E  & 22-Jul-24 & Preprint \\
    \bottomrule
    \end{tabular}
    }
  \label{tab:pfm_info}%
\end{table}%

\noindent
PathBench seeks to create a dynamically updatable evaluation platform for ongoing validation. As of March 2025, we have included 19 pathology foundation models that have publicly released their model weights, which can be categorized into three types: 1) Vision-only models, which are pretrained on pathological images only, 2) Vision-Language models, which leverage the paired data of pathological images and textual descriptions for pretraining, and 3) Multimodal models, which are enhanced by extra pathology-related modalities data. The details of PFMs information are summarized in Table ~\ref{tab:pfm_info}. Notably, we will continuously update the benchmark by incorporating additional PFMs once their weights are publicly released.
\\

\noindent
\textbf{(1) Vision-only PFM.} Prior to the era of foundation models, ResNet50~\cite{r50} pretrained on ImageNet was widely applied in the computational pathology community. Therefore, we include it as a baseline for validation. As a pioneering model in this field, CTransPath~\cite{ctranspath} first leveraged the MoCo-v3~\cite{mocov3} strategy for pre-training on over 30K publicly available H\&E slides. Building upon the iBOT framework, Phikon~\cite{phikon} empirically validated Vision Transformer's~\cite{vit} (ViT) ability to derive pan-cancer representation learning. As a groundbreaking work, UNI~\cite{uni} introduced the first model pre-trained on over 100,000 WSIs using DINOv2 framework~\cite{dinov2}. Additionally, it brought the issue of data contamination to the forefront, and established a robust evaluation benchmark across 34 representative pathological tasks to mitigate contamination risks. Recently, both Phikon and UNI have introduced their advanced versions, Phikon2~\cite{phikon2} and UNI2, by enlarging the pretraining data scale and IHC staining data. Virchow~\cite{virchow} first scaled pretraining data to the million-level while expanding the model architecture to ViT-H/14. Furthermore, Virchow2 pushed the boundary of PFM again by extending the pretraining data to over 3 million slides, resulting in the largest model in this field with 1.9B parameters (ViT-G/14). With the remarkable success of the DINOv2 framework in CPath community, Hibou-L~\cite{hibou},  H-Optimus-0 and H-Optimus-1 have consistently employed this architecture on over 1 million slides. Building upon these powerful PFMs,  GPFM~\cite{gpfm} incorporates the strengths of diverse PFMs by distilling their knowledge into a generalizable pathology foundation model. Beyond patch extractors pretraining,  Prov-GigaPath first broadened the scope of modelling into whole slides.
\\

\noindent
\textbf{(2) Vision-Language PFMs.} As the seminal vision-language foundation model in pathology, PLIP~\cite{plip} leveraged extensive pathological image-text pairs collected from Twitter, replies and LAION-5B~\cite{laion} to pretrain a model based on CLIP~\cite{clip} framework. CONCH~\cite{conch} crawled a substantial collection of image-text pairs from high-quality PubMed and educational sources, while additionally incorporating proprietary in-house datasets comprising paired pathology reports and electronic medical records for pretraining based on CoCa~\cite{coca} framework. Recently, CONCH has released the advanced version, CONCH1.5, with more pretraining data and a larger model. CHIEF designed a CLIP-like variant by leveraging the textual description of sites pretrained on over 60k slides. Furthermore, MUSK developed a multimodal transformer with unified masked modelling for vision-language modelling based on BEiT3 architecture.
\\

\noindent
\textbf{(2) Multimodal PFMs.} Beyond the modalities of vision and language, multimodal PFMs aim to incorporate more pathology-related modalities to enhance their capability of pathological image modelling. In this work, we involve mSTAR~\cite{mstar}, a whole-slide PFM, enhanced by pathology reports and gene expression data. Recently, more multimodal PFMs continue to emerge, such as THREADS~\cite{threads}, which is boosted by genomic and transcriptomic profiles. However, due to unavailable model weights, they have not yet been included at this stage. We will maintain ongoing monitoring and will dynamically incorporate them into our updatable evaluation platform once their models are released.
    \subsection*{3. Standardized Preprocessing}
    To ensure fair and reliable comparisons across heterogeneous WSI datasets,
    we established a standardized preprocessing pipeline addressing three key
    aspects: resolution variability, processing efficiency, and reproducibility.
    Our preprocessing approach consisted of the following steps:
    \begin{itemize}
        \item \textbf{Foreground extraction:} We exclusively analyzed foreground
            tissue patches, excluding background regions. Using consistent
            parameters for tissue detection ensured reproducibility across
            datasets. For slides with faint staining that challenged automated detection,
            we excluded them entirely rather than relying on manual annotations,
            thus maintaining methodological consistency.

        \item \textbf{Patch extraction:} All WSIs were processed at the base
            level (level 0), with patch dimensions scaled according to magnification:
            $512 \times 512$ pixels for 40$\times$ WSIs and $1024 \times 1024$ pixels
            for 80$\times$ WSIs. This configuration maintained a consistent tissue
            coverage of 0.25 $\mu$m$^{2}$/pixel across all samples.

        \item \textbf{Model-specific adaptations:} When implementing baseline
            models, we strictly followed their original specifications. For
            instance, with the MUSK model \cite{xiang2025vision}, we
            incorporated their recommended multiscale augmentation strategies as
            provided in the official implementation.
    \end{itemize}
    \subsection*{4. Evaluating Protocols}
    \subsubsection*{Paradigm}
Downstream tasks in this benchmark can be typically categorized into 2 types: 1) classification and 2) survival prediction. To fully demonstrate the capabilities of PFMs, we adopt the conventional two-stage multiple instance learning (MIL) paradigm for slide-level tasks. Following the mainstream evaluation strategy~\cite{uni, virchow}, ABMIL~\cite{abmil} is employed as an MIL aggregator to merge all patch features of a WSI into a slide-level representation using attention-based weighting, which is trained from scratch for every downstream task given patch features extracted by various PFMs. The details of training hyperparameters can be found in Table ~\ref{tab:hyper_param}, which are kept consistently for different PFMs to guarantee fair comparison.
\subsubsection*{Data Split}
 Both internal and external validation are employed for robust and generalizable evaluation. To maintain statistically valid assessments, for classification, the dataset for each task is split into training, validation, and test sets in a 7:1:2 ratio on every internal cohort, with experiments repeated 10 times using different random seeds. All 10 models generated from the repeated training process are subsequently assessed on external datasets to validate generalizability. For survival prediction, we adopt 5-fold cross-validation repeated 3 times experiments to achieve reliable comparisons given different seeds, resulting in 15 times repeated runs for each dataset. Similarly, all models trained across every fold and every seed are used for inference on external datasets.
\subsubsection*{Statistical Analysis}
To rigorously evaluate whether the observed assessment results demonstrate statistically significant differences, we apply non-parametric 1000 times bootstrapping for every experimental run. As a result, 10,000 and 15,000 bootstrap replicates are employed under the aforementioned data splits to estimate 95\% confidence intervals (CI) for every classification and survival task, respectively. For the best-performing model on every task, we estimate if it has a statistical difference from every other model via the one-sided Wilcoxon signed-rank test~\cite{wilcoxon}, and \textit{P}-value is subsequently reported.
\subsubsection*{Metrics}
For classification tasks, we report the AUC and its 95\% CI, which is a common metric used in classification independent of the decision threshold choice and remains unaffected by variations in class imbalance. For survival tasks, Concordance Index (C-Index) is commonly adopted, which represents the probability that, when two individuals are randomly chosen, their predicted risks will be ranked correctly.

    % include sample.bib file
    \bibliography{sample.bib}

    \section*{Supplementary information}

    \begin{table}[h]
        \centering
        \caption{The in-house data used for the benchmark. }
\begin{tabular}{ccccc}
            \hline
            Organ   & Source & Cohort            & Case Number & Slides Number \\
            \hline
            Lung    & H1     & Metastatic-Cohort & 846         & 1,422         \\
            Lung    & H5     & Metastatic-Cohort & 493         & 493           \\
            Lung    & H6     & Metastatic-Cohort & 826         & 1,422         \\
            Stomach & H7     & Biopsy-Cohort     & 1,345       & 2,700         \\
            Stomach & H1     & Gastric-Cohort    & 403         & 404           \\
            Stomach & H4     & Gastric-Cohort    & 320         & 320           \\
            Stomach & H3     & Gastric-Cohort    & 260         & 260           \\
            Colorectal   & H8     & Colorectal-Cohort      & 622         & 2,779         \\
 Colorectal& H1& Colorectal-Cohort& 294&301\\
            Breast  & H9     & Breast-Cohort     & 418         & 421           \\
            Breast  & H2     & Breast-Cohort     &             2,045& 4,275\\
            Brain   & H1     & Glioma-Cohort     & 677         & 1,362         \\
            \hline
        \end{tabular}
        \label{tab:data_details}
    \end{table}

    \begin{table}[htbp]
        \centering
        \caption{Performance of the evaluated models on the classification of
        primary and metastatic lung cancer task. One internal dataset (H1) and two
        external datasets (H5 and H6) are used for evaluation. The mean AUC and
        the 95\% confidence interval are reported. }
        \label{tab:lung_cancer_primary_metastatic}
        \begin{tabular}{lccc}
            \toprule Model    & Internal (H1)                   & External (H5)                   & External (H6)                   \\
            \midrule ResNet50 & 0.8835 (0.8236-0.9329)          & 0.6497 (0.5824-0.7129)          & 0.6819 (0.6393-0.7225)          \\
            UNI               & 0.9785 (0.9579-0.9932)          & 0.8690 (0.8253-0.9082)          & 0.8968 (0.8601-0.9235)          \\
            UNI2              & 0.9805 (0.9579-0.9948)          & 0.8743 (0.8359-0.9116)          & 0.8932 (0.8635-0.9189)          \\
            CONCH             & 0.9763 (0.9524-0.9922)          & 0.8610 (0.8190-0.8990)          & 0.8744 (0.8434-0.9000)          \\
            CHIEF             & 0.9755 (0.9541-0.9914)          & 0.8372 (0.7942-0.8772)          & 0.8339 (0.8030-0.8630)          \\
            GPFM              & 0.9815 (0.9614-0.9949)          & 0.8595 (0.8209-0.8974)          & 0.8963 (0.8705-0.9204)          \\
            mSTAR             & 0.9815 (0.9599-0.9951)          & \textbf{0.8811 (0.8371-0.9166)} & 0.9034 (0.8674-0.9277)          \\
\bottomrule
        \end{tabular}
    \end{table}

    \begin{table}[htbp]
        \centering
        \caption{Performance of the evaluated models on the task of predicting
        primary site of lung cancer. One internal dataset (H1) and two external datasets
        (H5 and H6) are used for evaluation. The mean AUC and the 95\% confidence
        interval are reported. }
        \label{tab:lung_cancer_primary_site_Prediction}
        \begin{tabular}{lccc}
            \toprule Model    & Internal (H1)                   & External (H5)                   & External (H6)                   \\
            \midrule ResNet50 & 0.8610 (0.7998-0.9142)          & 0.6927 (0.6423-0.7377)          & 0.8162 (0.7704-0.8541)          \\
            UNI               & 0.9742 (0.9503-0.9903)          & 0.9417 (0.9153-0.9629)          & 0.9777 (0.9670-0.9866)          \\
            UNI2              & 0.9735 (0.9493-0.9904)          & \textbf{0.9555 (0.9350-0.9725)} & 0.9830 (0.9740-0.9896)          \\
            CONCH             & 0.9622 (0.9328-0.9850)          & 0.9379 (0.9141-0.9576)          & 0.9795 (0.9692-0.9877)          \\
            CHIEF             & 0.9421 (0.9022-0.9720)          & 0.9108 (0.8803-0.9373)          & 0.9478 (0.9271-0.9649)          \\
            GPFM              & 0.9711 (0.9470-0.9883)          & 0.9341 (0.9045-0.9581)          & 0.9742 (0.9606-0.9833)          \\
            mSTAR             & 0.9738 (0.9494-0.9902)          & 0.9469 (0.9210-0.9680)          & 0.9796 (0.9704-0.9875)          \\
\bottomrule
        \end{tabular}
    \end{table}

    \begin{table}[ht]
        \centering
        \caption{Performance of the evaluted models on the biomarker prediction
        tasks in the lung cancer dataset. The mean AUC and the 95\% confidence
        interval are reported.}
        \begin{tabular}{lcccc}
            \toprule Model    & C-Met (H1)                      & CK7 (H1)                        & TTF-1 (H1)                      & NapsinA (H1)                    \\
            \midrule ResNet50 & 0.5541 (0.4304-0.6782)          & 0.6724 (0.5484-0.7885)          & 0.8726 (0.7917-0.9401)          & 0.8868 (0.7848-0.9636)          \\
            UNI               & 0.6643 (0.5261-0.7870)          & 0.9248 (0.8438-0.9856)          & 0.9645 (0.9018-0.9957)          & 0.9426 (0.8693-0.9894)          \\
            UNI2              & 0.5617 (0.4269-0.6985)          & 0.9211 (0.8347-0.9836)          & \textbf{0.9960 (0.9796-1.0000)} & 0.9695 (0.9149-0.9987)          \\
            CONCH             & 0.5671 (0.4289-0.7150)          & 0.9103 (0.8282-0.9738)          & 0.9644 (0.9081-0.9982)          & 0.9083 (0.8250-0.9696)          \\
            CHIEF             & 0.5912 (0.4563-0.7170)          & 0.8260 (0.7185-0.9149)          & 0.9721 (0.9324-0.9960)          & 0.9437 (0.8753-0.9900)          \\
            GPFM              & 0.6300 (0.4923-0.7652)          & 0.9002 (0.8096-0.9682)          & 0.9844 (0.9514-1.0000)          & 0.9574 (0.9048-0.9918)          \\
            mSTAR             & 0.6633 (0.5283-0.7875)          & 0.9191 (0.8366-0.9796)          & 0.9790 (0.9409-0.9983)          & 0.9381 (0.8648-0.9863)          \\
\bottomrule
        \end{tabular}
    \end{table}

    \begin{table}[ht]
        \centering
        \caption{Performance of the evaluated models on the molecular subtyping
        task in the breast cancer dataset. The mean AUC and the 95\% confidence
        interval are reported. }
        \begin{tabular}{lcc}
            \toprule Model    & Internal (H2)                   & External (H9)                   \\
            \midrule ResNet50 & 0.8880 (0.8670-0.9077)          & 0.7363 (0.7001-0.7706)          \\
            UNI               & 0.9239 (0.9049-0.9411)          & 0.7996 (0.7592-0.8377)          \\
            UNI2              & 0.9313 (0.9148-0.9465)          & 0.8188 (0.7816-0.8554)          \\
            CONCH             & 0.8992 (0.8775-0.9196)          & 0.7622 (0.7292-0.7935)          \\
            CHIEF             & 0.9144 (0.8949-0.9327)          & 0.7690 (0.7342-0.8022)          \\
            GPFM              & 0.9241 (0.9055-0.9410)          & 0.7958 (0.7469-0.8323)          \\
            mSTAR             & 0.9225 (0.9017-0.9408)          & 0.7859 (0.7453-0.8340)          \\
\bottomrule
        \end{tabular}
    \end{table}

    \begin{table}[ht]
        \centering
        \caption{Performance of the evaluated models on the tasks of TNM N
        staging (N0 vs N+) of breast cancer. The mean AUC and the 95\% confidence
        interval are reported. }
        \label{tab:breast_cancer_tnm_N}
        \begin{tabular}{lcc}
            \toprule Model    & Internal (H2)                   & External (H9)                   \\
            \midrule ResNet50 & 0.4102 (0.2721-0.5560)          & 0.4696 (0.3390-0.6026)          \\
            UNI               & 0.7035 (0.5637-0.8317)          & \textbf{0.7191 (0.5766-0.8453)} \\
            UNI2              & 0.6865 (0.5372-0.8232)          & 0.6889 (0.5437-0.8275)          \\
            CONCH             & 0.6741 (0.5209-0.8107)          & 0.6884 (0.5491-0.8155)          \\
            CHIEF             & 0.7082 (0.5751-0.8333)          & 0.6349 (0.4821-0.7769)          \\
            GPFM              & 0.6292 (0.4225-0.7969)          & 0.6804 (0.5390-0.8134)          \\
            mSTAR             & 0.6885 (0.5410-0.8249)          & 0.7008 (0.5630-0.8259)          \\
\bottomrule
        \end{tabular}
    \end{table}

    \begin{table}[ht]
        \centering
        \caption{Performance of the evaluated models on the biomarker prediction
        tasks including ER, PR and HER2 in the breast cancer dataset. The mean
        AUC and the 95\% confidence interval are reported. }
        \begin{tabular}{lccc}
            \toprule Model    & ER (H2)                         & PR (H2)                         & HER2 (H2)                       \\
            \midrule ResNet50 & 0.8236 (0.7728-0.8697)          & 0.8619 (0.8095-0.9078)          & 0.7145 (0.6520-0.7732)          \\
            UNI               & 0.8952 (0.8571-0.9289)          & 0.8914 (0.8497-0.9288)          & 0.8176 (0.7563-0.8714)          \\
            UNI2              & 0.9105 (0.8740-0.9426)          & 0.8732 (0.8248-0.9182)          & 0.8162 (0.7612-0.8675)          \\
            CONCH             & 0.8850 (0.8432-0.9216)          & 0.8707 (0.8272-0.9096)          & 0.8041 (0.7453-0.8596)          \\
            CHIEF             & 0.8655 (0.8174-0.9089)          & 0.8808 (0.8375-0.9201)          & 0.8031 (0.7458-0.8554)          \\
            GPFM              & 0.8944 (0.8533-0.9306)          & 0.8866 (0.8393-0.9265)          & 0.8358 (0.7826-0.8841)          \\
            mSTAR             & 0.8964 (0.8561-0.9318)          & 0.8847 (0.8378-0.9257)          & 0.8188 (0.7584-0.8716)          \\
\bottomrule
        \end{tabular}
    \end{table}

    \begin{table}[ht]
        \centering
        \caption{Performance of the evaluated models on the tasks of biomark
        prediction tasks inlcuding AR and CK5, and the pTNM staging task in the breast
        cancer dataset. The mean AUC and the 95\% confidence interval are reported.
        }
        \begin{tabular}{lccc}
            \toprule Model    & AR (H2)                         & CK5 (H2)                        & pTNM                            \\
            \midrule ResNet50 & 0.7115 (0.6389-0.7812)          & 0.7618 (0.6479-0.8552)          & 0.5438 (0.4330-0.6525)          \\
            UNI               & 0.7292 (0.6549-0.7987)          & 0.8300 (0.7521-0.8986)          & 0.5743 (0.4631-0.6843)          \\
            UNI2              & 0.7242 (0.6438-0.7982)          & \textbf{0.8481 (0.7718-0.9124)} & 0.5326 (0.4210-0.6467)          \\
            CONCH             & 0.7345 (0.6501-0.8089)          & 0.8076 (0.7120-0.8888)          & 0.5560 (0.4489-0.6708)          \\
            CHIEF             & 0.7218 (0.6518-0.7892)          & 0.7865 (0.6968-0.8646)          & 0.5642 (0.4502-0.6787)          \\
            GPFM              & \textbf{0.7447 (0.6764-0.8097)} & 0.8189 (0.7368-0.8915)          & 0.5876 (0.4790-0.6923)          \\
            mSTAR             & 0.7402 (0.6696-0.8077)          & 0.8322 (0.7526-0.9018)          & 0.5686 (0.4559-0.6818)          \\
\bottomrule
        \end{tabular}
    \end{table}

    \begin{table}[htbp]
        \centering
        \caption{Performance of the evaluated models on the overall survival
        prediction and disease-free survival prediction tasks in the breast
        cancer dataset. The mean C-Index and the 95\% confidence interval are
        reported. }
        \label{tab:breast_cancer_survival}
        \begin{tabular}{lcc}
            \toprule Model & OS (H2)                         & DFS (H2)                        \\
            \midrule CHIEF & 0.6796 (0.4861-0.8408)          & 0.6690 (0.4705-0.8495)          \\
            CONCH          & 0.6641 (0.4733-0.8307)          & 0.6352 (0.4617-0.7930)          \\
            CONCH1.5       & 0.6762 (0.4743-0.8446)          & 0.6461 (0.4664-0.8076)          \\
            CTransPath     & \textbf{0.6809 (0.4828-0.8489)} & 0.6449 (0.4618-0.8133)          \\
            Prov-GigaPath  & 0.6335 (0.4196-0.8281)          & 0.6482 (0.4390-0.8417)          \\
            GPFM           & 0.6523 (0.4532-0.8238)          & 0.6475 (0.4318-0.8488)          \\
            H-Optimus-0    & 0.6410 (0.4008-0.8315)          & 0.6093 (0.4068-0.8060)          \\
\bottomrule
        \end{tabular}
    \end{table}

    \begin{table}[htbp]
        \centering
        \caption{Performance of the evaluated models on the gastric biopsy
        dataset (Biopsy-Cohort). The evaluted three tasks are: detection of
        Intestinal Metaplasia, classification of Normal or Abnormal, and
        prediction of Autoimmune chronic gastritis with Helicobacter pylori (ACGxHP).
        The mean AUC and the 95\% confidence interval are reported. }
        \begin{tabular}{lccc}
            \toprule Model    & Intestinal Metaplasia           & Normal/Abnormal                 & ACGxHP                          \\
            \midrule ResNet50 & 0.9354 (0.8899-0.9686)          & 0.8704 (0.8319-0.9052)          & 0.7888 (0.7005-0.8555)          \\
            UNI               & 0.9421 (0.8940-0.9713)          & 0.9187 (0.8820-0.9454)          & 0.8574 (0.7811-0.9124)          \\
            UNI2              & 0.9555 (0.9245-0.9775)          & 0.9118 (0.8763-0.9437)          & 0.8510 (0.7893-0.9047)          \\
            CONCH             & 0.9535 (0.9196-0.9755)          & 0.8987 (0.8576-0.9299)          & 0.8578 (0.7867-0.9163)          \\
            CHIEF             & 0.9364 (0.8817-0.9673)          & 0.9143 (0.8480-0.9449)          & 0.8385 (0.7809-0.8894)          \\
            GPFM              & 0.9311 (0.8470-0.9734)          & 0.9134 (0.8798-0.9414)          & \textbf{0.8645 (0.8036-0.9123)} \\
            mSTAR             & 0.9465 (0.8869-0.9736)          & 0.9226 (0.8897-0.9480)          & 0.8447 (0.7663-0.9065)          \\
\bottomrule
        \end{tabular}
        \label{tab:gastric_biopsy_IM_NA_ACGxHP}
    \end{table}

    \begin{table}[htbp]
        \centering
        \caption{Performance of the evaluated models on the gastric biopsy
        dataset (Biopsy-Cohort). The evaluted three tasks are detection of Helicobacter
        pylori-associated chronic gastritis (HPACG), classification of gastric
        polyps, and classification of gastric ulcers. The mean AUC and the 95\% confidence
        interval are reported. }
        \begin{tabular}{lccc}
            \toprule Model    & HPACG                           & Polyp                           & Ulcer                           \\
            \midrule ResNet50 & 0.9081 (0.8398-0.9562)          & 0.8136 (0.7057-0.9053)          & 0.7401 (0.6192-0.8388)          \\
            UNI               & 0.9538 (0.9077-0.9810)          & 0.9457 (0.8785-0.9860)          & 0.8366 (0.7336-0.9241)          \\
            UNI2              & 0.9520 (0.9010-0.9802)          & 0.9444 (0.8877-0.9824)          & 0.8433 (0.7352-0.9285)          \\
            CONCH             & 0.9591 (0.9306-0.9790)          & 0.8990 (0.8003-0.9637)          & 0.8372 (0.7341-0.9248)          \\
            CHIEF             & 0.9413 (0.9034-0.9708)          & 0.9242 (0.8619-0.9713)          & 0.8133 (0.6297-0.9113)          \\
            GPFM              & 0.9440 (0.9003-0.9745)          & 0.9254 (0.8428-0.9800)          & 0.8590 (0.7633-0.9348)          \\
            mSTAR             & 0.9518 (0.9085-0.9786)          & 0.9472 (0.8880-0.9858)          & 0.8524 (0.7590-0.9313)          \\
\bottomrule
        \end{tabular}
        \label{tab:gastric_biopsy_HPACG_Polyp_Ulcer}
    \end{table}

    \begin{table}[htbp]
        \centering
        \caption{Performance of the evaluated models on the gastric grade
        Assesment task. One internal dataset (H1) and two external datasets (H3
        and H4) are used for evaluation. The mean AUC and the 95\% confidence interval
        are reported. }
        \label{tab:gastric_grade}
        \begin{tabular}{lccc}
            \toprule Model    & Internal (H1)                   & External (H3)                   & External (H4)                   \\
            \midrule ResNet50 & 0.6693 (0.4992-0.8235)          & 0.7336 (0.6193-0.8283)          & 0.7217 (0.5854-0.8292)          \\
            UNI               & 0.8502 (0.7371-0.9405)          & 0.8268 (0.7559-0.8897)          & 0.8451 (0.7827-0.8986)          \\
            UNI2              & \textbf{0.8834 (0.7894-0.9543)} & 0.8200 (0.7468-0.8834)          & 0.8232 (0.7490-0.8866)          \\
            CONCH             & 0.8377 (0.7312-0.9283)          & 0.8438 (0.7809-0.9009)          & \textbf{0.8626 (0.8053-0.9130)} \\
            CHIEF             & 0.7878 (0.6579-0.9007)          & 0.8479 (0.7790-0.9037)          & 0.8218 (0.7549-0.8819)          \\
            GPFM              & 0.8492 (0.7413-0.9397)          & 0.8170 (0.7496-0.8797)          & 0.8345 (0.7756-0.8867)          \\
            mSTAR             & 0.8562 (0.7538-0.9412)          & 0.8311 (0.7619-0.8914)          & 0.8453 (0.7855-0.8974)          \\
\bottomrule
        \end{tabular}
    \end{table}

    \begin{table}[htbp]
        \centering
        \caption{Performance of the evaluated models on the gastric Lauren
        subtyping. One internal dataset (H1) and two external datasets (H3 and
        H4) are used for evaluation. The mean AUC and the 95\% confidence interval
        are reported. }
        \label{tab:gastric_lauren}
        \begin{tabular}{lccc}
            \toprule Model    & Internal (H1)                   & External (H3)                   & External (H4)                   \\
            \midrule ResNet50 & 0.5692 (0.4582-0.6698)          & 0.5947 (0.5159-0.6713)          & 0.5582 (0.5075-0.6089)          \\
            UNI               & 0.7902 (0.7037-0.8737)          & 0.7843 (0.7041-0.8637)          & 0.7759 (0.7096-0.8332)          \\
            UNI2              & 0.7839 (0.6976-0.8657)          & 0.7897 (0.7315-0.8517)          & 0.7780 (0.7205-0.8252)          \\
            CONCH             & 0.8016 (0.7184-0.8784)          & 0.7990 (0.7341-0.8605)          & \textbf{0.7937 (0.7497-0.8360)} \\
            CHIEF             & 0.7775 (0.6925-0.8571)          & 0.7924 (0.7297-0.8511)          & 0.7911 (0.7453-0.8357)          \\
            GPFM              & 0.7783 (0.6914-0.8610)          & 0.7705 (0.7056-0.8481)          & 0.7842 (0.7394-0.8283)          \\
            mSTAR             & 0.7785 (0.6936-0.8588)          & 0.7945 (0.7235-0.8637)          & 0.7698 (0.6986-0.8340)          \\
\bottomrule
        \end{tabular}
    \end{table}

    \begin{table}[htbp]
        \centering
        \caption{Performance of the evaluated models on the gastric pathological
        subtyping task. One internal dataset (H1) and two external datasets (H3
        and H4) are used for evaluation. The mean AUC and the 95\% confidence interval
        are reported. }
        \label{tab:gastric_pathological_subtyping}
        \begin{tabular}{lccc}
            \toprule Model    & Internal (H1)                   & External (H3)                   & External (H4)                   \\
            \midrule ResNet50 & 0.6028 (0.5014-0.7084)          & 0.5312 (0.4525-0.6028)          & 0.5061 (0.4568-0.5533)          \\
            UNI               & 0.8079 (0.6975-0.8939)          & 0.6350 (0.5569-0.7168)          & 0.6531 (0.5784-0.7314)          \\
            UNI2              & 0.7928 (0.7000-0.8770)          & 0.6808 (0.5956-0.7640)          & 0.6691 (0.5887-0.7452)          \\
            CONCH             & 0.8023 (0.7120-0.8823)          & 0.6225 (0.5476-0.7384)          & 0.6250 (0.5627-0.7762)          \\
            CHIEF             & 0.7426 (0.6348-0.8375)          & 0.5879 (0.5011-0.6706)          & 0.5543 (0.4933-0.6197)          \\
            GPFM              & 0.7908 (0.6904-0.8777)          & 0.6855 (0.5964-0.7704)          & 0.6840 (0.5922-0.7759)          \\
            mSTAR             & 0.8057 (0.6913-0.8972)          & 0.6277 (0.5501-0.7108)          & 0.6438 (0.5802-0.7139)          \\
\bottomrule
        \end{tabular}
    \end{table}

    \begin{table}[htbp]
        \centering
        \caption{Performance of the evaluated models on the gastric Perineural
        Invasion task. One internal dataset (H1) and two external datasets (H3
        and H4) are used for evaluation. The mean AUC and the 95\% confidence interval
        are reported. }
        \label{tab:gastric_Perineural_Invasion}
        \begin{tabular}{lccc}
            \toprule Model    & Internal (H1)                   & External (H3)                   & External (H4)                   \\
            \midrule ResNet50 & 0.8385 (0.7293-0.9320)          & 0.7781 (0.7003-0.8477)          & 0.6937 (0.6315-0.7519)          \\
            UNI               & 0.9269 (0.8467-0.9863)          & 0.8457 (0.7863-0.8974)          & 0.7559 (0.6974-0.8098)          \\
            UNI2              & 0.9191 (0.8372-0.9800)          & 0.8461 (0.7853-0.8993)          & 0.7586 (0.7012-0.8136)          \\
            CONCH             & 0.9303 (0.8482-0.9886)          & 0.8457 (0.7870-0.8974)          & 0.7588 (0.7036-0.8109)          \\
            CHIEF             & 0.9118 (0.8310-0.9733)          & 0.8382 (0.7676-0.8975)          & 0.7463 (0.6837-0.8035)          \\
            GPFM              & 0.9284 (0.8519-0.9825)          & 0.8493 (0.7921-0.9007)          & 0.7632 (0.7070-0.8143)          \\
            mSTAR             & 0.9282 (0.8487-0.9876)          & 0.8456 (0.7885-0.8972)          & 0.7523 (0.6961-0.8059)          \\
\bottomrule
        \end{tabular}
    \end{table}

    \begin{table}[htbp]
        \centering
        \caption{Performance of the evaluated models on the gastric Vascular
        Invasion task. One internal dataset (H1) and two external datasets (H3
        and H4) are used for evaluation. The mean AUC and the 95\% confidence interval
        are reported. }
        \label{tab:gastric_Vascular_Invasion}
        \begin{tabular}{lccc}
            \toprule Model    & Internal (H1)                   & External (H3)                   & External (H4)                   \\
            \midrule ResNet50 & 0.7549 (0.6353-0.8619)          & 0.6593 (0.5826-0.7337)          & 0.6693 (0.5930-0.7353)          \\
            UNI               & 0.7475 (0.6287-0.8578)          & 0.7099 (0.6299-0.7833)          & 0.6879 (0.6124-0.7495)          \\
            UNI2              & 0.7817 (0.6701-0.8825)          & 0.7044 (0.6199-0.7831)          & 0.6966 (0.6282-0.7569)          \\
            CONCH             & 0.7680 (0.6551-0.8705)          & 0.7085 (0.6254-0.7845)          & 0.7142 (0.6487-0.7764)          \\
            CHIEF             & 0.7663 (0.6528-0.8703)          & 0.7182 (0.6443-0.7883)          & 0.7065 (0.6472-0.7633)          \\
            GPFM              & 0.7715 (0.6492-0.8786)          & 0.7345 (0.6596-0.8035)          & 0.7120 (0.6499-0.7701)          \\
            mSTAR             & 0.7642 (0.6468-0.8697)          & 0.7226 (0.6435-0.7959)          & 0.6952 (0.6330-0.7561)          \\
\bottomrule
        \end{tabular}
    \end{table}

    \begin{table}[htbp]
        \centering
        \caption{Performance of the evaluated models on the gastric biomarker
        detection task based on the Gastric-Cohort from center H1. The mean AUC and
        the 95\% confidence interval are reported. }
        \label{tab:gastric_biomarker}
        \begin{tabular}{lcc}
            \toprule Model    & S-100                           & HER-2                           \\
            \midrule ResNet50 & 0.7483 (0.6007-0.8776)          & 0.4919 (0.3634-0.6189)          \\
            UNI               & 0.8254 (0.6687-0.9433)          & 0.5727 (0.4242-0.7178)          \\
            UNI2              & 0.8347 (0.6177-0.9618)          & 0.5607 (0.4309-0.6869)          \\
            CONCH             & 0.8477 (0.7216-0.9449)          & 0.5317 (0.3975-0.6660)          \\
            CHIEF             & 0.7976 (0.6375-0.9231)          & 0.5357 (0.4105-0.6574)          \\
            GPFM              & 0.8135 (0.6667-0.9347)          & 0.5945 (0.4617-0.7213)          \\
            mSTAR             & 0.8335 (0.6988-0.9422)          & 0.5625 (0.3878-0.7089)          \\
\bottomrule
        \end{tabular}
    \end{table}

    \begin{table}[htbp]
        \centering
        \caption{Performance of the evaluated models on the gastric TNM-N task. One
        internal dataset (H1) and two external datasets (H3 and H4) are used for
        evaluation. The mean AUC and the 95\% confidence interval are reported. }
        \label{tab:gastric_TNM-N}
        \begin{tabular}{lccc}
            \toprule Model    & Internal (H1)                   & External (H3)                   & External (H4)                   \\
            \midrule ResNet50 & 0.7481 (0.6312-0.8510)          & 0.6572 (0.5784-0.7336)          & 0.6337 (0.5706-0.6963)          \\
            UNI               & 0.7820 (0.6485-0.8887)          & 0.7383 (0.6656-0.8080)          & 0.6927 (0.6318-0.7515)          \\
            UNI2              & 0.7908 (0.6786-0.8870)          & \textbf{0.7665 (0.6964-0.8322)} & 0.6899 (0.6096-0.7579)          \\
            CONCH             & 0.7831 (0.6701-0.8831)          & 0.7529 (0.6809-0.8198)          & 0.6887 (0.6291-0.7477)          \\
            CHIEF             & 0.7940 (0.6879-0.8856)          & 0.7503 (0.6790-0.8177)          & 0.6881 (0.6292-0.7455)          \\
            GPFM              & 0.7848 (0.6637-0.8819)          & 0.7476 (0.6726-0.8157)          & 0.6955 (0.6197-0.7560)          \\
            mSTAR             & 0.7916 (0.6726-0.8894)          & 0.7428 (0.6696-0.8122)          & 0.6957 (0.6368-0.7526)          \\
\bottomrule
        \end{tabular}
    \end{table}

    \begin{table}[htbp]
        \centering
        \caption{Performance of the evaluated models on the gastric TNM-T task. One
        internal dataset (H1) and two external datasets (H3 and H4) are used for
        evaluation. The mean AUC and the 95\% confidence interval are reported. }
        \label{tab:gastric_TNM-T}
        \begin{tabular}{lccc}
            \toprule Model    & Internal (H1)                   & External (H3)                   & External (H4)                   \\
            \midrule ResNet50 & 0.7183 (0.6298-0.7955)          & 0.6401 (0.5823-0.6963)          & 0.6708 (0.6258-0.7158)          \\
            UNI               & 0.7831 (0.7104-0.8504)          & 0.6763 (0.6353-0.7177)          & 0.7429 (0.7064-0.7801)          \\
            UNI2              & 0.8062 (0.7331-0.8724)          & 0.6369 (0.6066-0.6687)          & 0.7371 (0.6986-0.7764)          \\
            CONCH             & 0.8176 (0.7510-0.8779)          & 0.6741 (0.6363-0.7131)          & 0.7513 (0.7158-0.7862)          \\
            CHIEF             & 0.7631 (0.6866-0.8390)          & 0.6333 (0.5997-0.6679)          & 0.6944 (0.6551-0.7316)          \\
            GPFM              & 0.7894 (0.7197-0.8558)          & 0.6572 (0.6199-0.6963)          & 0.7351 (0.6968-0.7718)          \\
            mSTAR             & 0.7843 (0.7121-0.8520)          & 0.6713 (0.6324-0.7169)          & 0.7412 (0.7036-0.7789)          \\
\bottomrule
        \end{tabular}
    \end{table}

    \begin{table}[htbp]
        \centering
        \caption{Performance of the evaluated models on the gastric overall
        survival (OS) and disease free survival (DFS) tasks. The 5-fold cross-validation
        is performed. The mean C-Index and the 95\% confidence interval are reported.}
        \label{tab:gastric_OS_DFS}
        \begin{tabular}{lcc}
            \toprule Model & OS (H3)                         & DFS (H3)                        \\
            \midrule CHIEF & 0.6343 (0.5183-0.7338)          & 0.7164 (0.5248-0.8343)          \\
            CONCH          & 0.6481 (0.5468-0.7403)          & 0.7202 (0.5377-0.8315)          \\
            CONCH1.5       & 0.6636 (0.5630-0.7570)          & \textbf{0.7329 (0.5594-0.8374)} \\
            CTransPath     & 0.6372 (0.5327-0.7357)          & 0.7171 (0.5344-0.8377)          \\
            Prov-GigaPath  & 0.6400 (0.5176-0.7494)          & 0.7272 (0.5266-0.8387)          \\
            GPFM           & 0.6403 (0.5213-0.7447)          & 0.7180 (0.4973-0.8432)          \\
            H-Optimus-0    & 0.6518 (0.5222-0.7578)          & 0.7216 (0.5108-0.8431)          \\
\bottomrule
        \end{tabular}
    \end{table}

    \begin{table}[htbp]
        \centering
        \caption{Performance of the evaluated models on the glioma IDH mutation,
        pathological subtype classification and WHO grading tasks on the Glioma-Cohort
        from center H1. The mean AUC and the 95\% confidence interval are reported.
        }
        \label{tab:brain_cancer}
        \begin{tabular}{lccc}
            \toprule Model    & IDH Mutation                    & Pathological Subtype            & WHO Grading                     \\
            \midrule ResNet50 & 0.8308 (0.7553-0.8966)          & 0.8587 (0.8056-0.9081)          & 0.7689 (0.7006-0.8360)          \\
            UNI               & 0.8905 (0.8169-0.9501)          & 0.9067 (0.8411-0.9542)          & 0.8535 (0.7936-0.9066)          \\
            UNI2              & 0.8869 (0.8194-0.9446)          & \textbf{0.9469 (0.9086-0.9746)} & \textbf{0.8724 (0.8203-0.9195)} \\
            CONCH             & 0.8673 (0.7958-0.9282)          & 0.8943 (0.8424-0.9400)          & 0.8271 (0.7756-0.8766)          \\
            CHIEF             & 0.8265 (0.7420-0.9038)          & 0.8600 (0.7804-0.9155)          & 0.8193 (0.7645-0.8725)          \\
            GPFM              & 0.8886 (0.8183-0.9462)          & 0.9255 (0.8720-0.9667)          & 0.8422 (0.7821-0.8954)          \\
            mSTAR             & 0.8848 (0.8118-0.9452)          & 0.9125 (0.8511-0.9569)          & 0.8561 (0.8031-0.9047)          \\
\bottomrule
        \end{tabular}
    \end{table}

    \begin{table}[htbp]
        \centering
        \caption{Performance of the evaluated models on the classification of
        lymph node staging (N0 vs N+), tumor invasion depth (T1+T2 vs T3+T4, and
        T1 vs T2 vs T3 vs T4) tasks on the colorectal cancer dataset. The mean AUC
        and the 95\% confidence interval are reported. }
        \label{tab:colon_cancer_tnm}
        \begin{tabular}{lccc}
            \toprule Model    & N staging (H8)                  & T staging (4 class, H8)& T staging (2 class, H8)\\
            \midrule ResNet50 & 0.8553 (0.7826-0.9166)          & 0.7797 (0.6815-0.8637)          & 0.8556 (0.7330-0.9519)          \\
            UNI               & 0.8961 (0.8310-0.9500)          & 0.8455 (0.7650-0.9091)          & 0.9052 (0.8028-0.9798)          \\
            UNI2              & 0.8818 (0.8145-0.9391)          & 0.8633 (0.7913-0.9215)          & 0.9062 (0.8026-0.9832)          \\
            CONCH             & 0.8832 (0.8113-0.9420)          & 0.8466 (0.7575-0.9155)          & 0.8868 (0.7624-0.9751)          \\
            CHIEF             & 0.9006 (0.8386-0.9511)          & 0.8481 (0.7729-0.9113)          & 0.8681 (0.7453-0.9621)          \\
            GPFM              & 0.8901 (0.8260-0.9448)          & 0.8542 (0.7786-0.9189)          & 0.9152 (0.8247-0.9789)          \\
            mSTAR             & 0.8970 (0.8342-0.9497)          & 0.8561 (0.7897-0.9155)          & 0.9084 (0.8080-0.9801)          \\
\bottomrule
        \end{tabular}
    \end{table}

    \begin{table}[htbp]
        \centering
        \caption{Performance of the evaluated models on the classification of
        TNM staging task and consensus molecular subtyping task in the colorectal
        cancer. The mean AUC and the 95\% confidence interval are reported. }
        \label{tab:colon_cancer_tnm_cms}
        \begin{tabular}{lcc}
            \toprule Model    & TNM Staging                     & Consensus Molecular Subtyping   \\
            \midrule ResNet50 & 0.8724 (0.8207-0.9181)          & 0.5919 (0.5120-0.6726)          \\
            UNI               & 0.9076 (0.8641-0.9441)          & 0.7567 (0.6788-0.8307)          \\
            UNI2              & 0.9050 (0.8625-0.9413)          & 0.7735 (0.7004-0.8419)          \\
            CONCH             & 0.9096 (0.8633-0.9491)          & 0.7344 (0.6607-0.8042)          \\
            CHIEF             & 0.9197 (0.8828-0.9519)          & 0.7559 (0.6627-0.8300)          \\
            GPFM              & \textbf{0.9272 (0.8894-0.9587)} & 0.7565 (0.6847-0.8251)          \\
            mSTAR             & 0.9095 (0.8682-0.9448)          & 0.7547 (0.6767-0.8311)          \\
\bottomrule
        \end{tabular}
    \end{table}

    \begin{table}[htbp]
        \centering
        \caption{Performance of the evaluated models on the tasks of overall
        survival prediction, disease-free survival prediction and disease-specific
        survival prediction in the colorectal cancer. The mean C-Index and the 95\% confidence
        interval are reported. }
        \label{tab:colon_cancer_survival}
        \begin{tabular}{lccc}
            \toprule Model & OS (H8)                         & DFS (H8)                        & DSS (H1)                        \\
            \midrule CHIEF & 0.7482 (0.6549-0.8319)          & 0.7372 (0.6062-0.8450)          & 0.7035 (0.4491-0.9212)          \\
            CONCH          & 0.7472 (0.6472-0.8374)          & 0.7197 (0.6068-0.8363)          & 0.7256 (0.4554-0.9346)          \\
            CONCH1.5       & 0.7522 (0.6505-0.8418)          & 0.7188 (0.5985-0.8441)          & 0.7437 (0.4918-0.9326)          \\
            CTransPath     & \textbf{0.7570 (0.6685-0.8367)} & 0.7373 (0.6141-0.8478)          & 0.7311 (0.4654-0.9316)          \\
            Prov-GigaPath  & 0.7318 (0.6382-0.8159)          & 0.7319 (0.6103-0.8687)          & 0.7176 (0.4868-0.9194)          \\
            GPFM           & 0.7370 (0.6357-0.8293)          & 0.7321 (0.6058-0.8620)          & 0.7372 (0.4649-0.9375)          \\
            H-Optimus-0    & 0.7415 (0.6440-0.8302)          & 0.7362 (0.6165-0.8707)          & 0.7327 (0.5036-0.9256)          \\
\bottomrule
        \end{tabular}
    \end{table}

    \begin{table}[htbp]
        \centering
        \caption{Hyperparameters of ABMIL for classification and survival
        prediction in downstream tasks}
        \begin{tabular}{l|r|r}
            \toprule Hyperparameter & Classification & Survival \\
            \midrule Hidden dim     & 512            & 512      \\
            Dropout                 & 0.25           & 0.25     \\
            Batch size              & 1              & 1        \\
            Epochs                  & 50             & 30       \\
            Optimizer               & Adam           & Adam     \\
            Learning rate           & 2.00E-04       & 2.00E-04 \\
            Scheduler               & Cosine         & Cosine   \\
            Weight decay            & 1.00E-05       & 1.00E-05 \\
            Loss                    & Cross-entropy  & NLL      \\
            \bottomrule
        \end{tabular}%
        \label{tab:hyper_param}%
    \end{table}%
\end{document}